\documentclass{article}
\usepackage{bbm}
\usepackage{microtype}
\usepackage{enumitem}
\usepackage{graphicx}
\usepackage{subfigure}
\usepackage{booktabs} 
\usepackage{subcaption} 

\usepackage{multirow}
\usepackage{makecell}

\usepackage{hyperref}
\usepackage{multicol}
\usepackage{float}

\usepackage[accepted]{icml2023}

\usepackage{amsmath}
\usepackage{amssymb}
\usepackage{mathtools}
\usepackage{amsthm}
\usepackage{subcaption} 
\usepackage{lipsum}

\usepackage[capitalize,noabbrev]{cleveref}

\theoremstyle{plain}

\theoremstyle{definition}

\theoremstyle{remark}

\usepackage[textsize=tiny]{todonotes}

\setcitestyle{numbers,square}

\icmltitlerunning{EBGAN-MDN: An Energy-Based Adversarial Framework for Multi-Modal Behavior Cloning}

\begin{document}

\twocolumn[
\icmltitle{EBGAN-MDN: An Energy-Based Adversarial Framework for Multi-Modal Behavior Cloning}




\icmlsetsymbol{equal}{*}

\begin{icmlauthorlist}
\icmlauthor{Yixiao Li}{}
\icmlauthor{Julia Barth}{}
\icmlauthor{Thomas Kiefer}{}
\icmlauthor{Ahmad Fraij}{}
\end{icmlauthorlist}




\vskip 0.3in
]




\begin{abstract}
Multi-modal behavior cloning faces significant challenges due to mode averaging and mode collapse, where traditional models fail to capture diverse input-output mappings. This problem is critical in applications like robotics, where modeling multiple valid actions ensures both performance and safety. We propose EBGAN-MDN, a framework that integrates energy-based models, Mixture Density Networks (MDNs), and adversarial training. By leveraging a modified InfoNCE loss and an energy-enforced MDN loss, EBGAN-MDN effectively addresses these challenges. Experiments on synthetic and robotic benchmarks demonstrate superior performance, establishing EBGAN-MDN as a effective and efficient solution for multi-modal learning tasks.
\end{abstract}

\section{Introduction}

In many real-world tasks, the relationship between inputs and outputs is often ambiguous or multi-modal, creating challenges for traditional supervised learning models and behavior cloning (BC). These models assume unimodal mappings and optimize deterministic loss functions, such as the mean squared error (MSE), leading to averaged predictions that are suboptimal or nonsensical. In robotics, for example, this can result in unsafe behavior. Therefore, effectively addressing multi-modal mappings is critical in domains, where modeling diverse outcomes ensures both performance and safety. However, current solutions often struggle due to computational inefficiencies or limitations in capturing multi-modality. Filtering strategies address this challenge by removing conflicting samples but rely on explicit mode identification \cite{zamora2022learning}. Energy-based models (EBMs), such as Implicit Behavior Cloning (IBC) \cite{florence2022implicit}, effectively represent multi-modal distributions using energy landscapes. However, IBC uses iterative optimization during inference which makes it computationally prohibitive. Generative Adversarial Imitation Learning (GAIL) \cite{NIPS2016_cc7e2b87} and generative adversarial networks (GANs) \cite{NIPS2014_5ca3e9b1,mirza2014conditional} offer computational efficiency but tend to suffer from issues such as mode collapse and interpolation between valid outputs, as highlighted in \cite{ke2020imitationlearningfdivergenceminimization}. Attempts to combine GANs with EBMs, such as Zhao et al. \cite{zhao2017energybased}, used an objective that maps to total variation, which exacerbates mode collapse \cite{arjovsky2017wasserstein}.

\begin{figure}[h]
    \centering
    \begin{minipage}{0.2\textwidth}
        \centering
        \includegraphics[width=\textwidth]{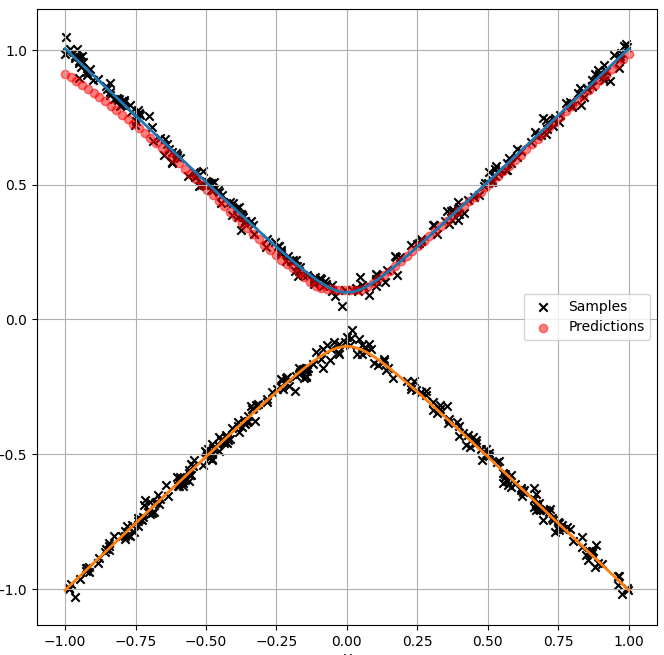}
        \text{(a) Mode collapse}
        
    \end{minipage}%
    \begin{minipage}{0.2\textwidth}
        \centering
        \includegraphics[width=\textwidth]{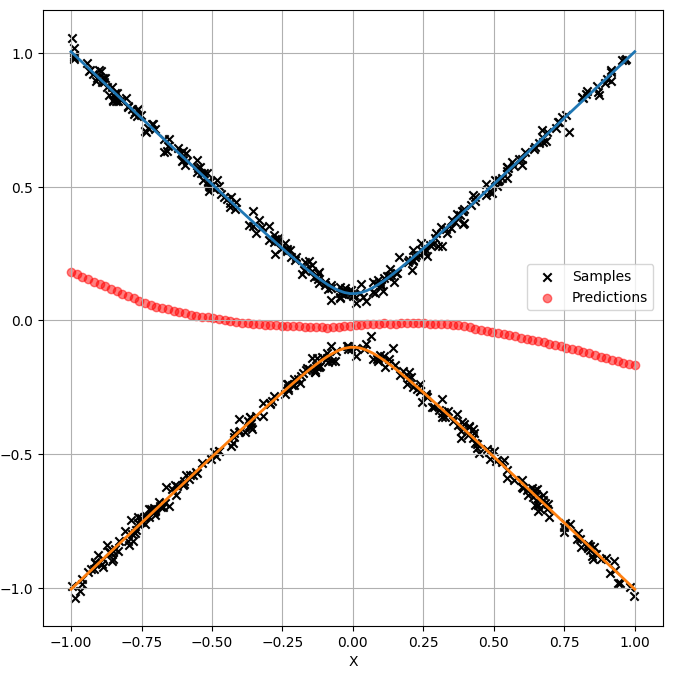}
        \text{(b) Mode averaging}

    \end{minipage}
    \caption{Comparison of mode collapse and mode averaging. (a) Mode collapse in a Vanilla cGAN model, which fails to capture diverse modes in the data. (b) Mode averaging using the Explicit BC model which produces unrealistic samples}
    \label{fig:mode_avg}
\end{figure}
To address these limitations, we propose \textbf{Energy-Based Generative Adversarial Networks with Mixture Density Networks (EBGAN-MDN)}, a framework that integrates energy-based models, GANs, and MDNs to effectively model multi-modal relationships. In EBGAN-MDN, an energy model evaluates input-output pairs, assigning lower energy to plausible mappings, while the generator minimizes this energy to produce valid and diverse outputs. By leveraging an MDN, the generator explicitly captures multi-modal distributions and 1-to-$k$ mappings for each input. Additionally, a modified InfoNCE loss refines the energy landscape, preventing mode collapse seen in traditional GANs and offering greater computational efficiency compared to iterative methods like IBC.

In summary, our contributions are as follows: \vspace{-5pt} 
\begin{itemize}[itemsep=-2pt] 
\item A novel fusion of energy-based adversarial frameworks with MDNs, enabling efficient modeling of multi-modal distributions. 
\item The introduction of an InfoNCE-based loss function and a energy-enforced MDN loss function. 
\item Validation of EBGAN-MDN on synthetic and real-world tasks, demonstrating superior performance in mode coverage, sample quality, and scalability compared to state-of-the-art baselines. \end{itemize}

\section{Models and Methods}

\subsection{The EBGAN-MDN}
\begin{figure}[h!]
    \centering
    \includegraphics[width=\linewidth]{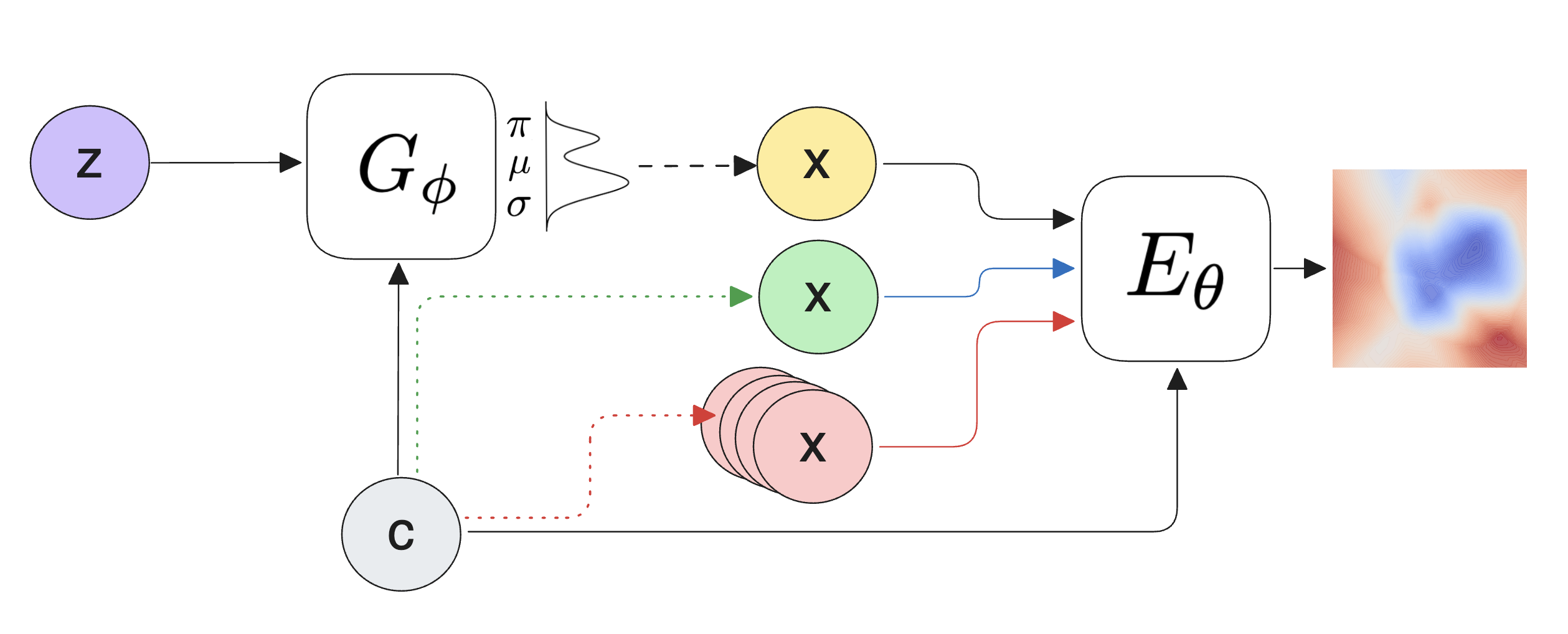}
    \caption{EBGAN-MDN architecture: The generator $G_\phi$ models multi-modal distributions by outputting Gaussian mixture parameters ($\pi, \mu, \sigma$) from latent $z$ and condition $c$. The energy model $E_\theta$ evaluates the plausibility of input-output pairs by assigning low energy to plausible pairs (green) and high energy to implausible ones (red), producing an energy landscape (low energy in blue, high energy in red).}
    \label{fig:Descriptive_image}
\end{figure}

Our proposed framework, \textbf{EBGAN-MDN}, combines an \textit{Energy-Based Model} with a \textit{Mixture Density Network} in an adversarial setup, which is illustrated in Figure \ref{fig:Descriptive_image}. This section describes the key components of the model and the corresponding loss functions as well as the algorithmic setup.

\subsubsection{Model Components}
\paragraph{Generator: Mixture Density Network}
In contrast to classic cGANs, we replace the typical MLP-based generator with a MDN. Our generator \( G_\phi: \mathbb{R}^{d_c + d_z} \rightarrow \mathbb{R}^{I \times (1 + d_{x} + d_{x} \times d_{x})} \) parameterizes a Gaussian Mixture Model (GMM) representing the conditional distribution \( p(x \mid c) \), where $d_x$ is the dimension of $x$. For a given latent input \( z \) and condition \(c\), the generator outputs the mixture weights (\(\pi\)), means (\(\mu\)), and covariances (\(\Sigma\)) for \( I \) Gaussians. Sampling from the GMM allows the generator to produce conflicting outputs, accessible through the different modes.
\paragraph{Energy Model}
The energy model \( E_\theta: \mathbb{R}^{d_c + d_x} \to \mathbb{R} \) assigns scalar energy values to input-output pairs \( (c, x) \). A lower energy value indicates a higher plausibility of the pair, guiding the generator \( G_\phi \) to produce outputs \( x \) that are diverse and consistent with the data distribution. This framework allows the energy model to evaluate the compatibility between \( c \) and \( x \), ensuring that the generator learns valid mappings \( (c \to x) \).


\subsubsection{Objectives}

\paragraph{The Energy Model Loss}  
The energy model aims to learn a landscape that assigns low energy to valid input-output pairs \((c, x)\) while penalizing invalid pairs with higher energy. This aligns with the contrastive InfoNCE objective proposed in \cite{florence2022implicit}. We adapted it to include synthetic samples generated by the generator for joint optimization of the energy model and generator and propose the following loss function:
\begin{align*}
&\mathcal{L}_{E_\theta}(X, z, c) = E_\theta(c, x) \\
&- \log \Big(e^{-E_\theta(c, x)} + \sum_{j=1}^{N} e^{-E_\theta(c, x_j)} + \alpha_t e^{-E_\theta(c, G_\phi(z \mid c))} \Big),
\end{align*}
where \(c\) is the condition, \(z\) is sampled from the latent space, and \(X = \{x, x_1, \ldots, x_N\}\) represents the data distribution samples. Here, \(x\) is the positive sample of a valid mapping (\(c \rightarrow x\)), \(\{x_j\}_{j=1}^N\) are negative samples of invalid mappings (\(c \nrightarrow x_j\)), and \(G_\phi(z \mid c)\) is a sample generated by the generator. The generator term ensures the model evaluates synthetic samples, pushing the generator to approximate valid outputs while distinguishing them from invalid pairs. We propose the optional use of a dynamic scaling factor \(\alpha_t\) which modulates the generator term's contribution, treating the sample as strong negative in the early stage and later decreasing its influence. For details on the loss design, we refer to Appendix \ref{sec:app_objective_funct} and for additional experiments with respect to $\alpha_t$ to Section \ref{Results_gen_config}.

\paragraph{The Generator Loss}  
The generator optimizes a dual objective: minimizing the energy of its samples and ensuring the generated GMM aligns with the true conditional distribution \(p(x \mid c)\) via a negative log-likelihood (NLL) term. The combined loss is given by:
\begin{align*}
\mathcal{L}_{G_\phi}(x, z, c) &= E_\theta(c, G_\phi(z \mid c)) \\
&- \log \sum_{i=1}^I \pi_i^{G_\phi(z \mid c)} \cdot f(x; \mu_i^{G_\phi(z \mid c)}, \Sigma_i^{G_\phi(z \mid c)}),
\end{align*}
where \(I\) is the number of Gaussians, and the generator produces the GMM parameters \(\{\pi_i, \mu_i, \Sigma_i\}_{i=1}^I\) conditioned on \(c\) and \(z\). This loss encourages the generator to minimize energy while accurately capturing the multimodal structure of the data. Further design details can be found in Appendix \ref{GenLoss_app}.

\subsubsection{Training Process}
The training process, outlined in Algorithm \ref{algorithm:training}, alternates between updating the energy model and the generator. For each batch, the energy model is updated multiple times using the energy-based loss, which is computed from real data pairs, negative samples, and generator outputs. This ensures that the energy model can distinguishes between valid and invalid pairs. Negative samples can be obtained by randomly selecting $x$ values from the data domain, ensuring that they do not have a valid mapping to $c$. The generator is then updated using a combined loss: an energy-based term that encourages outputs to receive low energy scores, and the NLL term that ensures accurate modeling of the conditional distribution.

\begin{algorithm}[h!]
\caption{Training EBGAN-MDN Model}
\begin{algorithmic}[1]
\STATE \textbf{Input:} Data loader $D$, Energy model $E_\theta$, Generator $G_\phi$, Optimizers $O_e$, $O_g$, Epochs $T$, EM Updates $T_E$, Negative samples $N$, Scaling factor $(\alpha_t)_t$
\FOR{t = 1 \textbf{to} $T$}
    \FOR{batch in $D$}
        \STATE Get $x, c$ from $D$
        \FOR{i = 1 \textbf{to} $T_E$}
            \STATE Sample $z$, generate $G_\phi(z \mid c)$
            \STATE Select $N$ negative samples $x_{\text{neg}}$
            \STATE $L_{E_\theta} = \mathcal{L}_{E_\theta}(c, x, x_{\text{neg}}, G_\phi(z \mid c), \alpha_t)$
            \STATE Update $E_\theta$ with $O_e$
        \ENDFOR
        \STATE Sample $z$, generate $G_\phi(z \mid c)$
        \STATE $L_{\text{g}} = E_\theta(c, G_\phi(z \mid c)) + \text{NLL}(G_\phi, x)$
        \STATE Update $G_\phi$ with $O_g$
    \ENDFOR
\ENDFOR
\end{algorithmic}
\label{algorithm:training}
\end{algorithm}
\subsection{Experimental design}

\paragraph{Baseline Models} We use Explicit-BC, cGAN \cite{mirza2014conditional}, MDN \cite{MDNs}, IBC \cite{florence2022implicit}, and our own EBGAN variant as baselines. Explicit-BC is implemented as a Vanilla MLP trained with MSE loss and MDN is benchmarked with the same number of Gaussians as our model to ensure fair comparison. IBC follows the implementation in \cite{florence2022implicit}, using an EBM with iterative inference. We additionally propose EBGAN, which follows our EBGAN-MDN approach but pairs an energy model with a neural generator instead of an MDN generator.

\paragraph{Benchmark 1: 2D Geometric Shapes} 
We benchmark the models with two synthetic experiments: the two-modal \textit{hyperbola} function \((\pm \sqrt{a^2 + b^2c^2})\) and the three-modal \textit{line} function \((\pm m\cdot c, 0)\). Metrics include KL divergence and Wasserstein distance to evaluate distribution alignment, as well as average modes captured and total mode coverage percentage to measure the ability to represent all modes. Detailed metric, model, and training implementation specifics are in Appendix ~\ref{Implementation_app}.
\begin{figure}[h!]
    \centering
    \includegraphics[width=0.5\textwidth]{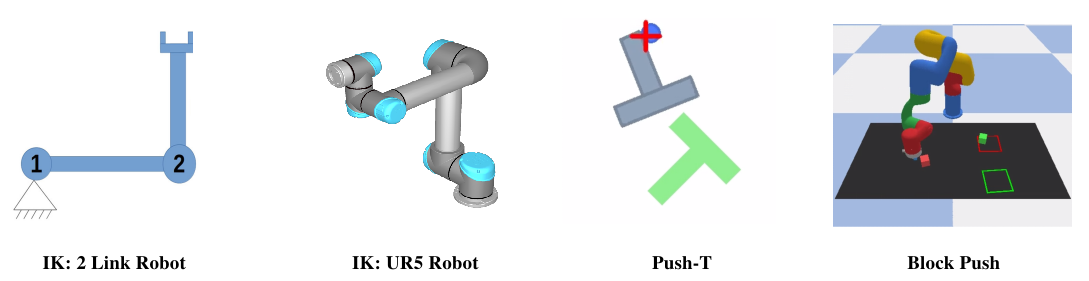} 
    \caption{Robot behaviour cloning tasks for evaluation}
    \label{fig:eval_tasks}
\end{figure}
\vspace{-1em}
\paragraph{Benchmark 2: Robotic Tasks} 
We evaluate EBGAN-MDN for BC
across a variety of robotic task \cite{florence2022implicit,chi2023diffusion}(Fig. \ref{fig:eval_tasks}) to compare its performance against the baseline models. The tasks include: \textit{IK 2-link Robot}, learning inverse kinematics for a 2-link arm with two valid configurations per target; \textit{IK UR5 Robot}, learning inverse kinematics for a 6-DOF UR5 arm with multiple valid joint configurations; \textit{Push-T}, pushing a T-shaped object to a target using multimodal strategies; and \textit{Block Push}, pushing two blocks to targets in varied orders and directions, introducing multimodality. The metrics include best and mean success rates, defined as the percentage of successful target completions. Full implementation details are in Appendix ~\ref{Implementation_app}.






\begin{figure}[htbp]
\vskip -0.05in
\begin{center}
\centerline{\includegraphics[width=\columnwidth]{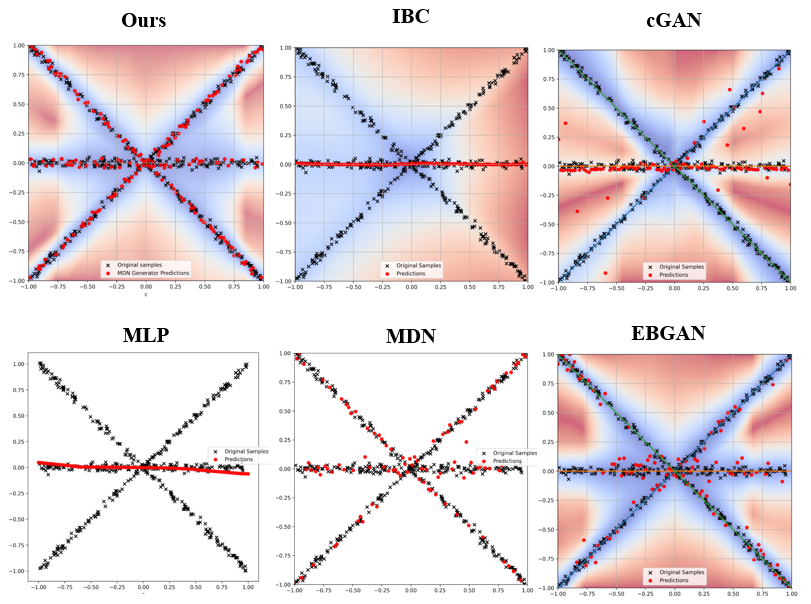}}
\caption{Visualization of model predictions and energy landscapes (or discriminator landscape) on the lines task. Red indicating low plausibility and blue indicating high plausibility.}
\label{ig:Full_comp_image_B1_H_H}
\end{center}
\vskip -0.2in
\end{figure}
\section{Results}\label{results}
\subsection{Benchmark 1: 2D Geometric Shapes}
Table~\ref{model-comparison-combined} show that EBGAN-MDN outperforms all baselines in all metrics in the hyperbola experiment. Explicit-BC fails to handle multi-modal tasks capturing zero modes by suffing from mode averaging while EBGAN persistently collapses to one single mode (TMC$=0$). cGAN shows instability (high variance) and partial mode collapse while beeing able to capture around $60\%$ of times all two modes. IBC performs reasonably well but is less consistent than EBGAN-MDN and presents with more outlier. MDN closely follows EBGAN-MDN but lacks the robustness of EBGAN-MDN. Similar results can be found in the Line experiment which highlights that MDN-based models, particularly EBGAN-MDN, effectively capture the three-modal structure, as evident in Figure ~\ref{ig:Full_comp_image_B1_H_H}, whereas all other models suffer from severe mode collapse or poor mode coverage. The full results can be found in Appendix \ref{Benchmark_1}.
\begin{table}[h!]
\vskip 0.15in
\begin{center}
\begin{small}
\begin{sc}
\begin{tabular}{lcccc}
\toprule
Model        & KL     & W       & TMC (\%)            & AMD   \\
\midrule
Explicit BC  & $14.54$         & $0.0281$          & $0.00 \pm 0.00$           & $0.032$ \\
MDN          & $2.41$          & $0.0080$          & $98.80 \pm 1.17$          & $1.988$ \\
cGAN         & $4.67$          & $0.0119$          & $65.00 \pm 37.64$         & $1.642$ \\
IBC          & $3.83$          & $0.0098$          & $76.80 \pm 3.31$          & $1.768$ \\
\midrule
EBGAN        & $6.92$          & $0.0140$          & $0.00 \pm 0.00$           & $0.990$ \\
EBGAN-MDN    & $\mathbf{2.11}$ & $\mathbf{0.0076}$ & $\mathbf{99.40 \pm 0.80}$ & $\mathbf{1.994}$ \\
\bottomrule
\end{tabular}
\end{sc}
\end{small}
\end{center}
\caption{Comparison of Models on Benchmark 1: Hyperbola. Metrics include KL Divergence (KL), Wasserstein Distance (W), Total Mode Coverage (\%), and Average Modes Captured (AMD).}
\label{model-comparison-combined}
\vskip -0.1in
\end{table}

\subsection{Benchmark 2: Robotic Tasks}
\begin{table}[htbp]
    \vskip 0.15in
    \setlength{\tabcolsep}{3.5pt} 
    \begin{center}
    \begin{small}
    \begin{sc}
    \begin{tabular}{lcccc}
    \toprule
    \multirow{2}{*}{} & \multicolumn{1}{c}{IK 2 link} & \multicolumn{1}{c}{IK UR5} & \multicolumn{1}{c}{Push-T} & \multicolumn{1}{c}{Block Push} \\ 
    \midrule
    Explicit BC  & 0.042 & 0.039 & 0.619 & 0.512 \\
    IBC       & 0.685  & 0.682 & 0.844 & 0.589  \\
    MDN       & 0.715 & 0.728 & 0.665 & 0.574 \\ 
    cGAN       & 0.883 & 0.826 & 0.551 & 0.387 \\ \midrule
    EBGAN & \textbf{0.921} & \textbf{0.920} & 0.845 & 0.469 \\
    EBGAN-MDN & 0.893 & 0.832 & \textbf{0.871} & \textbf{0.754} \\  \bottomrule
    \end{tabular}
    \end{sc}
    \end{small}
    \end{center}
    \caption{Success rate with each averaged across 50 different robot task initial conditions.}
    \label{tab:performance_comparison}
    \vskip -0.1in
\end{table}

As shown in Table \ref{tab:performance_comparison}, EBGAN achieves the highest performance in most tasks, with notable improvements in inverse kinematics tasks (IK 2-link and IK UR5), where it outperforms all other methods. For the multimodal tasks, Push-T and Block Push, EBGAN-MDN exhibits strong results, particularly in Push-T, where it achieves competitive success rates. While EBGAN-MDN slightly underperforms EBGAN in IK tasks, it excels in Block Push, achieving the highest success rates compared to all methods, showcasing its effectiveness in handling complex multimodal planning.

\section{Discussion}
\textbf{Comparison with IBC, cGAN, and MDN.}   EBGAN-MDN draws inspiration from IBC, cGAN, and MDN. Unlike IBC, which requires optimization during inference, it uses a generator for direct mapping, boosting inference speed. Compared to cGAN, EBGAN-MDN prevents mode collapse by incorporating an MDN for diverse, consistent outputs. Unlike MDN, it benefits from adversarial training, which enhances sample quality. Results (Section \ref{results}) show that EBGAN-MDN outperforms IBC, GAN, and MDN in diversity, consistency, and generative performance.

\textbf{Sharpness.} To address the potential limitation of MDNs in handling sharp edges due to their Gaussian assumptions, we tested different noise models for the MDN generator (e.g., diagonal, isotropic, fixed noise levels, and Laplace) on Benchmark 1. Fixed noise levels achieved high mode coverage ($\geq 99\%$) and near-optimal KL Divergence for the hyperbola benchmark but lacked flexibility in more general scenarios. Laplacian noise performed similarly well, with its heavier tails better capturing outliers near each mode, maintaining both high mode coverage and tight distribution alignment. More details can be found in the Appendix \ref{noise}.

\subsection{Strengths}
\textbf{Avoiding Mode Collapse with EBGAN-MDN.}  The MDN generator overcomes the limitations of traditional generators, which often collapse to a single mode despite latent space encoding theoretically allowing 1-to-$k$ mappings. By generating distributions instead of direct samples, MDNs capture data diversity and produce plausible outputs. As shown in the Line experiment, only MDN-based models successfully captured the task's three-modal structure, achieving notably better accuracy than explicit BC, cGANs, and EBGANs (see \ref{ig:Full_comp_image_B1_H_H}).

\textbf{Energy Landscape Guidance.}  EBMs provide more informative guidance than standard discriminator through continuous energy scores instead of binary learning signal. This is reflected in the results: EBGAN-MDN surpasses standard MDNs, achieving a higher TMC (\(99.4\%\)) in the hyperbola task and better success rates in robotic tasks. Furthermore, we could observe in robotics tasks, that EBGAN-MDN and EBGANs outperforms cGAN by a large margin, demonstrating the energy model’s effectiveness.

\subsection{Limitations}
While EBGAN-MDN demonstrates strong performance in multi-modal BC, it has some limitations. The number of Gaussians in the MDN must be approximately known a priori, as incorrect assumptions can lead to underfitting, impacting performance. Additionally, the MDN framework is not well-suited for high-dimensional data like images, limiting its applicability to vision tasks. Finally, the adversarial training, combined with the MDN's parameter estimation, increases training computational overhead.

\section{Summary}
In this work, we proposed EBGAN-MDN, a framework combining energy-based models, MDNs, and adversarial training to address multi-modal BC. By leveraging a modified InfoNCE loss and an energy-enforced MDN loss, EBGAN-MDN effectively resolves mode averaging and mode collapse. Experiments on synthetic and robotic benchmarks demonstrated superior performance compared to methods like explicit BC, cGAN, IBC, and MDNs, establishing EBGAN-MDN as an effective and efficient solution to these challenges. The code for it, along with all relevant experiments, is publicly available on GitHub\footnote{\href{https://github.com/zipping-suger/MultiModLearn}{\texttt{github.com/zipping-suger/MultiModLearn}}}.

\clearpage
\bibliography{main}
\bibliographystyle{ieeetr}

\clearpage
\appendix
\clearpage
\onecolumn
\section{Background and Related Work}

\paragraph{Energy Models for Behavioral Cloning}

Implicit Behavior Cloning (IBC) reformulates behavioral cloning as a conditional energy-based modeling (EBM) problem, addressing the limitations of traditional explicit models. Instead of directly mapping observations to actions, IBC represents the policy as an optimization problem where the optimal action minimizes an energy function conditioned on the observation. In the context of imitation learning, IBC improves performance on tasks with high-dimensional action spaces, visual observations, and real-world environments. IBC uses the InfoNCE (Information Noise Contrastive Estimation) loss function to train the energy-based model. Given that \( E_{\theta}(x_i, y_i) \) is the energy function scoring the observed action, \( \tilde{y}_{ij} \) represents negative counter-examples, and \( N_{\text{neg}} \) is the number of negative samples per batch \cite{florence2022implicit}, the loss is defined as:
\[
 -\sum_{i=1}^{N} \log \left( \frac{\exp(-E_\theta(x_i, y_i))}{\exp(-E_\theta(x_i, y_i)) + \sum_{j=1}^{N_{\text{neg}}} \exp(-E_\theta(x_i, \tilde{y}_{ij}))} \right)
\]
A primary limitation when comparing to explicit models is that energy-based models typically require more compute, both in training and inference, leading to longer inference times.

\paragraph{(Conditional) GAN}
Generative Adversarial Networks (GANs) consist of two competing networks: a generator \(G(z)\) that maps random noise \(z\) to synthetic data, and a discriminator \(D(x)\) that distinguishes real data \(x\) from generated samples. They are trained in a minimax game where the generator tries to fool the discriminator, while the discriminator improves at identifying fake data \cite{NIPS2014_5ca3e9b1}. Conditional GANs (cGANs) extend GANs by conditioning both the generator and discriminator on additional information \(c\), enabling controlled data generation. The generator produces samples \(G(z \mid c)\) tailored to the condition, while the discriminator verifies both the data and its conditioning. The objective function \cite{mirza2014conditional} of cGAN is:

\[
\min_{G} \max_{D} \mathbb{E}_{x,c \sim p_\text{data}} [\log D(x \mid c)] + \mathbb{E}_{z \sim p_z} [\log(1 - D(G(z \mid c)))].
\]

cGANs are essential for behavior cloning (BC) tasks because they allow the generator to produce outputs \(G(z \mid c)\) that are conditioned on c. This is crucial in BC scenarios where the model needs to generate outputs x that correspond to specific conditions c. Thus, cGANs facilitate controlled data generation and enable the model in BC tasks to learn mapping between input conditions c and multiple target labels $x_1, \dots, x_k$.

\paragraph{Wasserstein GAN}
WGAN (Wasserstein GAN) is a variant of GANs that uses the Wasserstein distance (Earth Mover's Distance) as a metric for comparing distributions, which helps to address training instability commonly found in traditional GANs \cite{arjovsky2017wasserstein}. The WGAN was one of the most popular GAN variants to successfully mitigate mode collapse.

\paragraph{Energy-Based GANs (EBGANs)} EBGANs introduce an energy-based perspective to the GAN framework by replacing the discriminator with an energy function, which assigns scalar values to input-output pairs \cite{zhao2017energybased}. The energy function attributes low energy to plausible data samples (close to the data manifold) and higher energy to implausible ones. The generator is then trained to produce samples in low-energy regions, while the energy-based discriminator is optimized to differentiate real samples from generated ones using a margin-based loss. EBGANs provide flexibility in the architecture of the discriminator and improve training stability over traditional GANs by allowing the use of loss functionals like reconstruction errors.

However, our proposed EBGAN-MDN framework differs fundamentally. While EBGANs typically employ neural networks or auto-encoders to compute the energy values, our approach integrates an MDN-based generator instead of the standard MLP or auto-encoder discriminator setup. This addition enables explicit multi-modal modeling, allowing the generator to learn and represent complex conditional distributions $p(x \mid c)$ while avoiding mode collapse. Furthermore, we use a different objective for the energy model which is close to the InfoNCE loss. It was proven that the loss function for EBGANs align with optimizing the total variation distance (TV) between the true and generated data \cite{arjovsky2017wasserstein}. Arjovsky et al. \cite{arjovsky2017wasserstein} showed that GANs trained with the TV distance are suffering from similar stability issues as the vanilla GAN making EBGAN suboptimal for mutli-modal applications.

\paragraph{Generative Adversarial Imitation Learning (GAIL)} GAIL is an extension of GANs designed specifically for imitation learning tasks, where the goal is to learn a policy that mimics expert behavior \cite{NIPS2016_cc7e2b87}. The generator in GAIL aims to produce actions that the discriminator cannot distinguish from expert actions, effectively learning a policy that imitates the expert’s behavior. This setup allows GAIL to learn from expert demonstrations without requiring explicit supervision. 


\paragraph{Mixture Density Networks (MDN) and Mixture Density GAN (MD-GAN)}
Mixture Density Networks (MDNs) are a class of probabilistic models that parameterize a Gaussian Mixture Model, producing outputs such as mixture weights, means, and variances. This allows MDNs to explicitly model multi-modal data distributions, capturing the inherent diversity of the data through sampling from the GMM's parameterized distribution \cite{MDNs}.

MD-GAN extends this approach by integrating an MDN into the generator of a GAN. Instead of producing direct samples, the generator outputs GMM parameters that define a probability distribution over possible outputs. Meanwhile, the discriminator organizes embeddings of real data into clusters represented as vertices of a simplex, effectively encouraging the generator to map diverse modes to these clusters. Unlike EBGAN-MDN, which directly uses an MDN generator to model multi-modality in the data, MD-GAN relies on the clustering behavior of the discriminator's embedding space to indirectly encourage mode diversity \cite{MDGAN}.

\subsection{Problem formulation}

\paragraph{Mode Collapse and Mode Averaging} Mode collapse occurs in GANs when the generator fails to capture the full diversity of the data distribution, instead producing samples from only a limited subset of the data's modes. This issue arises because the generator is optimized to fool the discriminator, which may be satisfied even when the generator only replicates a small portion of the data space\cite{MDGAN}. Figure \ref{fig:mode_avg} presents this mode collapse problem in which the GAN's generator is only able to capture the concave-up portion of the hyperbola, completely neglecting other data points. Mode averaging is also related to mode collapse. It is when the generator produces outputs that fall between the modes of the true data distribution rather than accurately representing any specific mode. This often results from the generator attempting to balance capturing multiple modes simultaneously, leading to unrealistic samples as seen in figure \ref{fig:mode_avg}. While mode collapse generates samples from only a few modes, mode averaging creates ambiguous samples that do not correspond to any true mode

\paragraph{Predicting Conflicting Samples}
In tasks involving conflicting samples, a single input $c$ corresponds to multiple valid target labels $\{x_1,\ldots,x_k\}$, a relationship known as a 1-to-$k$ mapping. Traditional supervised learning methods, such as regression, neural networks, or tree-based models, are inherently designed for deterministic 1-to-1 mappings. These models typically optimize loss functions like mean squared error or cross-entropy (CE), collapsing all possible outcomes into a single averaged prediction $\hat{y}$. We denote inputs as $c$ since they take the postion of the condition in a GAN framework.

Addressing this limitation requires models that can represent multi-valued functions and explicitly capture the multi-modal nature of the data. This can be achieved using optimization-based methods, such as those described in \cite{florence2022implicit}, which treat the solution as a set of optimizers (argmin) over the objective landscape. 
Alternatively, the model can learn the multi-modal distribution $p(x\mid c)$ instead of an explicit 1-to-$k$ mapping, enabling conflicting samples to be obtained by sampling from its $k$-modal structure.
Probabilistic models like Gaussian Mixture Models and Conditional Variational Autoencoders  explicitly parameterize $p(x\mid c)$, capturing distinct modes. Generative models, such as GANs, implicitly learn the distribution, generating samples via transformations of latent variables. Deterministic multi-output models, such as Mixture Density Networks, predict parameters of a mixture distribution (e.g., means, variances, weights) to model contrasting modes. However, while these models excel at learning distributions, they often struggle with effectively capturing multi-modal structures in practice.
\section{Deriving the objective functions} \label{sec:app_objective_funct}

\subsection{The energy model objective}

\subsubsection{Energy Models and their connection to probability distributions}
The fundamental connection between energy and probability is derived from the Gibbs distribution \cite{yann_EBM}:
\begin{align}
    P (X \mid C) = \frac{\exp{(-\beta E (X,C))}}{Z(C)}
\end{align}
where \(C\) and \(X\) are two random variables, $\beta$ is an arbitrary positive constant (like an inverse temperature), and $Z(C)$ is the partition function defined as:
\begin{align*}
    Z(C) = \int_{x \in \mathcal{X}} \exp{(-\beta E (x, C))} dx
\end{align*}
We will set $\beta$ to $1$. An EBM $E_\theta$ is used to estimate $E$. Depending on the how well $E_\theta$ is learned from the data, this serves as a good estimate of the probabilities \cite{yann_EBM}. 

\subsubsection{Mutual Information and InfoNCE for Energy-Based Models}

Mutual information (MI) quantifies the dependency between two random variables \(C\) (condition) and \(X\) (target). It is formally defined as:

\[
I(C; X) = \mathbb{E}_{(c, x) \sim p(c, x)} \left[\log \frac{p(c, x)}{p(c)p(x)}\right] = \mathbb{E}_{(c, x) \sim p(c, x)} \left[\log \frac{p(x \mid c)}{p(x)}\right] ,
\]

where \(p(c, x)\) represents the joint distribution, and \(p(c)\) and \(p(x)\) are the marginals of \(C\) and \(X\), respectively. Intuitively, \(I(C; X)\) measures how much knowing \(C\) reduces uncertainty about \(X\) and vice versa. In tasks involving conditional mappings, maximizing mutual information \(I(C; X)\) ensures the model effectively captures the meaningful relationships between inputs $C$ (which are conditions in the GAN notation) and outputs $X$ (which are input and outputs in the GAN notation). Mutual information quantifies the shared information between $C$ and $X$, providing a measure of how well the mapping between them is established. For instance, in the 2-Link robot arm problem, where $C$ represents the probability variable for the position in Cartesian space and $X$ corresponds to the probability variable of the joint angles required to reach that position, a high mutual information \(I(C; X)\) indicates a strong correspondence between the input conditions and the valid target outputs. Moreover, mutual information should be low between $C$ and invalid or unrelated outputs $X$. Thus, maximizing \(I(C; X)\) plays a critical role in learning accurate and robust condition-output relationships. \\

Directly maximizing \(I(C; X)\) is intractable due to the difficulty in estimating the marginal \(p(x)\). The InfoNCE objective is introduced to approximate MI \cite{oord2018representation}:
\begin{align}
    \mathcal{L}_{\text{InfoNCE}} = - \mathbb{E}_{X} \left[\log \frac{f(x,c)}{\sum_{j=1}^{N} f(x_j,c)}\right],
\end{align}
where $X = \{x_1, x_2\ldots x_N\}$ is a set of $N$ random samples containing $N-1$ negative samples and one positive sample $x$ (e.g.,  $ x := x_N$). These negative samples are drawn from the marginal distribution \(p(x)\), which are unrelated to \(c\) while the positive sample is drawn from the conditonal distribution $p(x \mid c)$. $f(x,c)$ is a density ratio which aims to perserve the mutual information between $x$ and $c$. It provides a lower bound to the mutual information:
\begin{align}
I(C; X) \geq log(N) - \mathcal{L}_{\text{InfoNCE}}.
\end{align}
which indicates that minimizing $\mathcal{L}_{\text{InfoNCE}}$ also maximizes a lower bound on mutual information \cite{oord2018representation}. \\

Florenc et al.~\cite{florence2022implicit} placed the energy model into the InfoNCE formulation and is proposing the following training loss for a training dataset $\{c_i,x_i\}^D_{i=1}$:
\begin{align}
\mathcal{L}_{\text{InfoNCE}} = \sum_{i=1}^D - \log \frac{\exp(-E_\theta(c_i, x_i))}{\exp(-E_\theta(c_i, x_i)) + \sum_{j=1}^{N-1} \exp(-E_\theta(c_i, x_i^j))},
\end{align}
where \(E_\theta(c, x)\) is an energy function $E_\theta: \mathbb{R}^{d_c + d_x}\rightarrow \mathbb{R}$ parameterized by \(\theta\) that measures the energy between the condition \(c\) and the target \(x\). This is the the negative log likelihood of 
\begin{align}
    p_\theta(x \mid c) \approx \frac{\exp(-E(c,x))}{Z_{X_i}(c)}.
\end{align}
where $Z_{X_i}(c)$ is the approximation of the normalization of the likelihood $Z(c)$ by $X_i = \{x_i^1, x_i^2\ldots x_i^N\}$ of the InfoNCE loss. This approach allows the model to sidestep direct computation of $p(x)$ while ensuring that the energy landscape is shaped according to the mappings between $c$ and $x$.\\

\subsubsection{InfoNCE and Contrastive Learning}
InfoNCE is a classic contrastive learning loss function. The goal of contrastive learning is to maximize mutual information (or a proxy for it as in our case) between the positive sample and the anchor, while minimizing similarities with negative samples. This requires contrasting the anchor with multiple negatives to sufficiently represent the broader data distribution. Contrastive learning operates and thereby perform its typical movements (pushing and pulling) in the space defined by the scoring function, which in our case is an energy model. This means the energy landscape is transformed using this loss function. In our framework, the terms "pulling" and "pushing" in the energy space describe how the optimization process reshapes the energy landscape to distinguish valid input-output pairs from invalid ones. The energy function $E_\theta (c,x)$ assigns a scalar energy value to each pair $(c, x)$, where lower energy corresponds to higher compatibility or plausibility of the relationship between $c$ (the condition or anchor) and $x$ (the true data or output). 

The denominator normalizes the energy over all positive and negative samples, effectively contrasting the valid pair \((c_i, x_i)\) against the negative samples \((c_i, x_i^j)\). The optimization process encourages the model to assign low energy to valid pairs and high energy to invalid ones, enabling the energy function to approximate the MI between \(C\) and \(X\). When we say valid pairs are "pulled" in the energy space, it means their energy values are reduced, effectively moving their predicted energy score toward a low-energy region. Conversely, invalid pairs are "pushed" in the energy space, which means their energy values are increased, moving them to a high-energy region.

The InfoNCE loss formalizes this behavior. The numerator $\exp(-E_\theta(c,x))$, prioritizes decreasing the energy of valid pairs, effectively pulling them toward the low-energy regions of the scalar field. The denominator includes contributions from both valid and invalid pairs, particularly terms like $\exp(-E_\theta(c,x^j))$ for negative samples. These terms encourage the energy model to assign higher energy to invalid pairs, pushing them to the high-energy regions. Together, these opposing forces reshape the energy function to clearly delineate valid and invalid pairs.

\subsubsection{Extending InfoNCE to Include the Generator Perspective}

The InfoNCE objective in its standard form focuses on contrasting positive pairs with negative samples from \(p(x)\). While effective for maximizing mutual information between $c$ and $x$, this formulation does not explicitly account for the role of a conditional generator \(G_\phi(z \mid c)\) in approximating $p(x \mid c)$. To address this, we extend the InfoNCE denominator to include generator-sampled pairs, providing additional information to the energy model about the evolving nature of the conditional distribution. The extended InfoNCE objective is:

\[
\mathcal{L}_{\text{EBM}}(\theta, \phi) = -  \mathbb{E}_{X} \left[
\log \frac{\exp(-E_\theta(c, x))}{\exp(-E_\theta(c, x)) + \sum_{j=1}^{M} \exp(-E_\theta(c, x_j)) + \mathbb{E}_{z \sim \mathcal{Z}} \left[\exp(-E_\theta(c, G_\phi(z \mid c)))\right]}
\right].
\]
In this formulation, the numerator, \(\exp(-E_\theta(c, x))\), represents the energy assigned to valid (positive) pairs. The denominator now contains three terms:
\begin{itemize}
    \item \(\exp(-E_\theta(c, x))\), the energy of the positive pair.
    \item \(\sum_{j=1}^{M} \exp(-E_\theta(c, x_j))\), the energy of negative samples drawn from \(p(x)\).
    \item \(\mathbb{E}_{z \sim \mathcal{Z}} \left[\exp(-E_\theta(c, G_\phi(z \mid c)))\right]\), the energy of samples generated by the generator \(G_\phi\) using the latent variable $z$
\end{itemize}

We can formulate the generalized energy model training loss as:
\[
\mathcal{L}_{\text{EBM}}(\theta, \phi) = -  \sum_{i=1}^D
\log \frac{\exp(-E_\theta(c_i, x_i))}{\exp(-E_\theta(c_i, x_i)) + \sum_{j=1}^{M} \exp(-E_\theta(c_i, x_i^j)) + \sum_{k=1}^{N}\exp(-E_\theta(c_i, G_\phi(z_k \mid c_i)))}
\]

\paragraph{Role of the Generator Term}  The inclusion of the generator term ensures that the energy model is exposed to samples from the generator’s distribution. This is particularly important because \(G_\phi(z \mid c)\) \(G_\phi\) evolves over the training to approximate $p(x \mid c)$. The generator term ensures that the energy model not only learns from the training data but also adapts to the evolving distribution of generated samples. Moreover, including the generator term in the denominator adds additional samples that help refine the approximation of $Z(c)$, providing a more comprehensive representation of the broader data distribution. We do not include generator samples in the numerator of the InfoNCE objective to ensure the energy model learns directly from the true data distribution. This approach prevents the risk of introducing noise from inaccurate generator outputs, particularly during early training, and maintains a clear separation between the energy model's role of capturing ground-truth relationships and the generator's task of approximating the conditional distribution $p(x \mid c)$. In addition, the generator initially produces invalid pairs \((c, G_\phi(z \mid c))\) that serve as effective negative samples. These generated samples help regularize the energy model by exposing it to plausible, but initially incorrect, predictions. 

\subsubsection{Mitigating the Bias of the Trained Generator on the Loss} 
As the generator improves, it begins producing samples that are closer to the true distribution \(p(x \mid c)\). These samples become indistinguishable from valid pairs, resulting in lower energy values for generated samples. By design, the denominator in the InfoNCE objective contrasts the positive pair (valid \((c, x)\)) with all negative samples, which includes those generated by the generator. This contrastive mechanism encourages the energy model to assign higher energy values to the samples in the denominator. While this is beneficial for incorrect generator samples, valid generator outputs could be inadvertently pushed toward higher energy regions. This dynamic interplay creates tension between the energy model and the generator, as the generator is simultaneously trained to minimize the energy of its outputs.

To address this tension, we propose \textbf{Dynamic Scaling}, a mechanism that gradually reduces the weight of valid generator samples in the denominator over the course of training. This approach allows the generator to initially contribute strongly to the contrastive signal when its samples are less valid, but diminishes its influence as the generator improves and begins producing more plausible outputs. The dynamic scaling factor, \(\alpha_t\), is defined as:
\[
\alpha_t = \max\{1 - \frac{t}{T}, a_{\min}\},
\]
where \(t\) is the current epoch, \(T\) is the total number of training epochs, and \(a_{\min}\) is a minimum scaling value (set to 0.1 in our experiments) to ensure that the generator’s influence is never completely eliminated. By incorporating \(\alpha_t\) into the denominator of the InfoNCE objective, we dynamically modulate the generator term as follows:
\[
\mathcal{L}_{\text{EBM}}(\theta, \phi) = -  \sum_{i=1}^D
\log \frac{\exp(-E_\theta(c_i, x_i))}{\exp(-E_\theta(c_i, x_i)) + \sum_{j=1}^{M} \exp(-E_\theta(c_i, x_i^j)) + \alpha_t\exp(-E_\theta(c_i, G_\phi(z_i \mid c_i)))}.
\]
here, \(\alpha_t\) ensures that the generator's contribution to the denominator decreases as the outputs align closer to the true distribution. Early in training, the generator term is treated as strongly negative, helping the energy model distinguish invalid samples. As training progresses, the reduced scaling factor prevents penalization of plausible generator outputs, allowing the energy model and generator to align effectively.

To further explore the impact of the generator term in the denominator, we conducted experiments with various loss configurations (Section~\ref{experiment_gen}), modifying the generator's contribution to the contrastive signal in the InfoNCE objective. The goal was to evaluate how these configurations influenced the balance between the generator and energy model during training. We recommend the use of this dynamic scaling factor to mitigate bias introduced by the generator in the energy model’s loss for stabilization. As shown in our experiments (results ~\ref{Results_gen_config}), this approach stabilizes training by reducing the tension between the generator and energy model. By gradually diminishing the generator’s influence, dynamic scaling allows the energy model to maintain a robust contrastive signal for invalid samples while preserving the plausibility of valid outputs. 

\subsection{The generator loss} \label{GenLoss_app}
\subsubsection{The objective and design of the generator loss}
The goal of the generator in EBGAN-MDN is to model the conditional distribution \(p(x \mid c)\), capturing the multimodal nature of the target distribution while aligning with the energy model's learned landscape. Unlike standard adversarial frameworks where the generator competes with the discriminator, the generator here complements the energy model by learning to generate samples that align with the true data distribution. This allows the generator to represent complex relationships between the condition \(c\) and the target \(x\), effectively approximating the underlying conditional mapping.

The training of the generator aims to achieve two key objectives: (1) minimizing the energy assigned by the energy model to the generator's outputs, and (2) ensuring the generator's outputs align with the Gaussian mixture model parameterization. These objectives are jointly addressed by a composite loss function that combines an energy-based term and a probabilistic term. The energy-based term ensures that the generator produces plausible outputs in accordance with the energy landscape, while the probabilistic term ensures the generator accurately captures the multimodal distribution through the GMM.

The generator loss is defined as:
\[
\mathcal{L}_{G_\phi}(x, z, c) = E_\theta(c, G_\phi(z \mid c)) - \log \sum_{i=1}^I \pi_i^{G_\phi(z \mid c)} \cdot f(x; \mu_i^{G_\phi(z \mid c)}, \sigma_i^{G_\phi(z \mid c)}).
\]

Here, \(E_\theta(c, G_\phi(z \mid c))\) represents the energy assigned by the energy model \(E_\theta\) to the condition-output pair \((c, G_\phi(z \mid c))\), where \(c\) is the condition and \(G_\phi(z \mid c)\) is the sample generated by the generator given the latent vector \(z\). Lower energy values indicate higher compatibility between \(c\) and \(G_\phi(z \mid c)\), encouraging the generator to produce plausible outputs.

The second term, is the negative log-likelihood of the target \(x\) under a Gaussian Mixture Model parameterized by the generator:
\[
\mathcal{L}_{\text{MDN}}(x,z,c) = -\log \sum_{i=1}^I \pi_i^{G_\phi(z \mid c)} \cdot f(x; \mu_i^{G_\phi(z \mid c)}, \sigma_i^{G_\phi(z \mid c)})
\]
\begin{itemize}
    \item \(I\) is the number of Gaussian components in the mixture.
    \item \(G_\phi(z \mid c)\) defines the parameter of the GMM. The parameters $\pi, \mu$, and $\sigma$ in this formula are retrieved from the generator output of $z$ and condition $c$. This results in the parameter set $\{\pi_i^{G_\phi(z \mid c)}, \mu_i^{G_\phi(z \mid c)}, \sigma_i^{G_\phi(z \mid c)}\}$.
    \item \(\pi_i^{G_\phi(z \mid c)}\) is the mixing coefficient for the \(i\)-th Gaussian component.
    \item \(f(x; \mu_i^{G_\phi(z \mid c)}, \sigma_i^{G_\phi(z \mid c)})\) is the probability density function (PDF) of a Gaussian distribution with mean \(\mu_i^{G_\phi(z \mid c)}\) and standard deviation \(\sigma_i^{G_\phi(z \mid c)}\), evaluated at \(x\).
\end{itemize}

\subsubsection{The working mechanism of the generator loss}
The first term, \(E_\theta(c, G_\phi(z \mid c))\), minimizes the energy assigned to the generator's outputs, effectively encouraging the generator to produce samples that the energy model deems plausible. The second term, \(-\log \sum_{i=1}^K \pi_i \cdot f(y; \mu_i, \sigma_i)\), represents the negative log-likelihood of the generator’s outputs under the GMM parameterization, ensuring that the generator learns a distribution that aligns with the multimodal nature of the data.

This loss aligns with the generator's goal because it directly penalizes deviations from the learned energy landscape while simultaneously enforcing probabilistic alignment with the target distribution. By minimizing the energy of the generated samples, the generator iteratively improves its outputs, producing samples that are closer to the true data distribution. Simultaneously, the GMM loss ensures that the generator captures the inherent multimodality of the target data, enabling it to model one-to-many mappings.

Traditional adversarial frameworks often suffer from mode collapse or mode averaging \cite{arjovsky2017wasserstein, ke2020imitationlearningfdivergenceminimization}, where the generator fails to represent the full diversity of the data. Similarly, directly minimizing the energy without incorporating the GMM loss would not enforce the multimodal nature of the distribution, leading to suboptimal representations (see EBGAN results in Results \ref{results}). Conversely, relying solely on the GMM loss without the energy term would disregard the broader context provided by the energy model, resulting in samples that might align with the data distribution but fail to capture the nuances of the energy landscape (see MDN results in Results \ref{results}). 

The combined loss function provides a principled approach to address these challenges. It integrates the strengths of energy-based learning and probabilistic modeling, ensuring that the generator captures both the diversity of the data and its alignment with the learned energy landscape. This dual optimization framework is crucial for effectively modeling complex, multimodal relationships in conditional mapping tasks.

\subsection{Combined Loss}

The training of our framework can be formulated as a minimax optimization problem. The objective combines the learning dynamics of the energy model \(E_\theta\) and the generator \(G_\phi\), ensuring that both components jointly optimize their roles in modeling the conditional distribution \(p(x \mid c)\).

\begin{align*}
    \min_{\phi} \max_{\theta} \mathbb{E}_{X} &\left[ 
    -\log \frac{e^{-E_\theta(c, x)}}{e^{-E_\theta(c, x)} + \sum_{j=1}^{N} e^{-E_\theta(c, x_j)} + \mathbb{E}_{z \sim \mathcal{Z}} \left[ e^{-E_\theta(c, G_\phi(z \mid c))} \right]} \right. \\
    &\quad + \left. \mathbb{E}_{z \sim \mathcal{Z}} \left[\log \sum_{i=1}^K \pi_i^{G_\phi(z \mid c)} \cdot f(x; \mu_i^{G_\phi(z \mid c)}, \sigma_i^{G_\phi(z \mid c)})\right] \right]
\end{align*}

The first term in the loss function corresponds to the InfoNCE objective for the energy model, which seeks to minimize the energy \(E_\theta(c, x)\) for valid input-target pairs \((c, x)\) while maximizing the energy for invalid pairs, including both negative samples from the dataset and samples generated by \(G_\phi\). This ensures that the energy model learns to distinguish between valid and invalid mappings in the energy space. By maximizing the first term with respect to \(\phi\), the energy of the generator's output pairs \((c, G_\phi(z \mid c))\) is minimized, aligning with the generator's objective of producing outputs that achieve low energy under the energy model. 

The second term corresponds to the negative log-likelihood under the generator's Gaussian Mixture Model representation. It enforces that the generator \(G_\phi\) models the true conditional distribution \(p(x \mid c)\) by minimizing the discrepancy between the generator's output distribution and the data distribution.

By combining these objectives, the minimax optimization aligns the energy model and the generator. The energy model learns to identify plausible mappings in the energy landscape, while the generator is trained to produce valid samples that align with both the data distribution and the energy model's learned landscape. This synergy ensures that the model handles multimodal distributions effectively and improves both plausibility and diversity in the generated outputs.

\section{Experiment Details}\label{Implementation_app}

\subsection{Benchmark 2D Geometric Shapes }\label{Benchmark_1}

\subsubsection{Evaluation Metrics} \label{Metrics}
The performance of the models is evaluated using four metrics, each designed to assess a different aspect of the models' ability to capture the underlying $m$-modal data distribution and conflicting samples. These metrics are: \textbf{total mode coverage}, \textbf{average modes captured}, \textbf{KL divergence}, and \textbf{Wasserstein distance}.

\paragraph{Total mode coverage (TMC)} We measure mode coverage of conflicting samples by the percentage of test inputs for which the model $G$ successfully predicts all true modes within a predefined tolerance \( \epsilon \). For each test condition \( c \), the true modes \( \{x_1, x_2, \ldots, x_m\} \) are computed. The model then produces \( K \) samples \( \{x^\text{G(c)}_1, x^\text{G(c)}_2, \ldots, x^\text{G(c)}_K\} \) for the same condition $c$. A generated value \( x^\text{G(c)}_i \) is considered to match a true mode \( x_j \) if 
\[
|x^\text{G(c)}_i - x_j| \leq \epsilon.
\]
Total mode coverage is then defined as the percentage of test inputs \( c \) for which all true modes \( \{x_1, x_2, \ldots, x_m\} \) are matched by at least one generated sample  \( \{x^\text{G(c)}_1, x^\text{G(c)}_2, \ldots, x^\text{G(c)}_K\} \):
\[
\text{Total mode Coverage (\%)} = \frac{\sum_{c \in \mathcal{C}_\text{test}} \mathbbm{1}\left\{\forall x_j \in \{x_1, x_2, \ldots, x_m\}, \exists x^\text{G(c)}_i : |x^\text{G(c)}_i - x_j| \leq \epsilon \right\}}{|\mathcal{C}_\text{test}|} \times 100,
\]
where \(|\mathcal{C}_\text{test}| \) is the test set.

\paragraph{Average modes captured (AMC)} The metric average modes captured provides a more nuanced evaluation by calculating the average number of true modes matched per test input. It is defined as:
\[
\text{Average Modes Captured} = \frac{1}{|\mathcal{C}_\text{test}|} \sum_{c \in \mathcal{C}_\text{test}} \sum_{x_j \in \{x_1, x_2, \ldots, x_m\}} \mathbbm{1}\left\{\exists x^\text{G(c)}_i \in  \{x^\text{G(c)}_1, x^\text{G(c)}_2, \ldots, x^\text{G(c)}_K\} : |x^\text{G(c)}_i - x_j| \leq \epsilon \right\},
\]
where \( m \) is the total number of true modes for each \( c \) and $x^\text{G(c)}_i$ are the samples produced by model $G$ with respect to $c$.

\paragraph{KL divergence (KL)} The KL divergence measures the difference between the true and generated output distributions. Let \( p(x) \) and \( q(x) \) represent the true and generated distributions, respectively, which are approximated using histograms with \( B \) bins. The KL divergence is computed as:
\[
D_\text{KL}(p \parallel q) = \sum_{b=1}^B p_b \log\left(\frac{p_b}{q_b}\right),
\]
where \( p_b \) and \( q_b \) are the normalized histogram densities in bin \( b \). To ensure numerical stability, a small constant \( \delta > 0 \) is added to both \( p_b \) and \( q_b \).

\paragraph{Wasserstein distance (WD)} The Wasserstein distance quantifies the cost of transforming the generated distribution \( q(x) \) into the true distribution \( p(x) \). For the same histogram approximations, the Wasserstein distance is given by:
\[
W(p, q) = \inf_{\gamma \in \Pi(p, q)} \int |x - x'| \, d\gamma(x, x'),
\]
where \( \Pi(p, q) \) is the set of all joint distributions with marginals \( p \) and \( q \). In practice, the Wasserstein distance is computed using the histograms of \( p(x) \) and \( q(x) \).

These metrics collectively evaluate the models' ability to (1) identify all true modes (\textbf{total mode coverage}), (2) partially capture modes (\textbf{average modes captured}), and (3) approximate the overall data distribution (\textbf{KL divergence} and \textbf{Wasserstein distance}). A lower KL Divergence and Wasserstein Distance indicate better distribution alignment, while higher Mode Coverage and Average Modes Covered reflect better mode diversity.

\subsubsection{Baseline Models}
In this series of experiments, we will benchmark six models: 
\begin{enumerate}
    \item \textbf{EBGAN-MDN (our method)}, a hybrid approach combining energy-based adversarial training with a mixture density network to explicitly model multi-modal distributions;
    \item \textbf{Explicit Behavioral Cloning (BC)}, a neural network trained with a mean squared error loss;
    \item \textbf{Mixture Density Networks (MDNs)}, which predict a probability distribution over possible outputs for each input (the generator of our framework);
    \item \textbf{Conditional GANs (cGANs)}, which generate \(y\)-values conditioned on \(x\);
    \item \textbf{Energy-based cGANs (EBGAN)}, combining an energy-based model with a generative adversarial component (which is not the model proposed in \cite{zhao2017energybased} but similar to our model with a standard neural network instead of a MDN); 
    \item \textbf{Implicit Behavior Cloning (IBC)}, which uses an energy-based model combined with an optimization to recover \(x\) by minimizing an energy function.
\end{enumerate}

We will use the metrics as defined in \ref{Metrics} including (1) \textbf{total mode coverage}, (2) \textbf{average modes captured}, (3) \textbf{KL divergence} and (4) \textbf{Wasserstein distance}. Accordingly, the number of true modes is $m=2$ and the test set is as described above ($\|\mathcal{C}_{\text{test}}\| = 100$). We set the number of samples 
$K=10$ per condition $c$ to ensure sufficient diversity in the generated outputs while maintaining computational efficiency during evaluation. The tolerance $\epsilon$ to $0.07$  is chosen to be smaller than the minimum distance between the two modes of the hyperbolic function ($\approx 0.201$), ensuring clear differentiation between conflicting samples. The number of bins $B$ is set to $50$. The configuration is used for both experiments \textit{hyperbola} and \textit{lines}. 

All models are trained five times to account for variability in training dynamics. The metrics are reported as the mean and standard deviation across these five runs. 

\subsubsection{Implementation Details}

The specific training hyperparameters for each model are summarized in Table~\ref{tab:hyperparameters}, while Table~\ref{tab:architectures} describes the architectural details.

\begin{table}[h!]
\centering
\begin{tabular}{|l|c|c|c|c|c|c|c|c|}
\hline
\textbf{Model} & \textbf{Gaussians} &  \textbf{Hidden Size} & \textbf{Epochs} & \textbf{BS} & \textbf{LR} & \textbf{LR-E/D} & \textbf{Neg. Sample} & \textbf{E/D-Runs} \\
\hline
EBGAN-MDN & 2 & 64 & 100 & 32 & 0.0005 & 0.001 & 32 & 5 \\
\hline
Explicit BC &  - & 64 & 100 & 32 & 0.001 & - & - & - \\
\hline
MDN & 2 &  64 & 100 & 32 &  0.001 & - & - & - \\
\hline
cGAN & - &  64 & 100 & 32 &  0.0002 & 0.002 & - & 5 \\
\hline
EBGAN & - &  64 & 100 & 32 & 0.0005 & 0.001 & 32 & 5\\
\hline
IBC & - &  64 & 100 & 32 & - & 0.001 & 64 & - \\
\hline
\end{tabular}
\caption{Hyperparameters for each model. BS is the Batch Size, LR and LR-E/D are the learning rates where LR-E/D captures learning rates of discriminators and Energy models and LR typically for any other used model and E/D-Runs are the number of inner EBM or Discriminator Updates in the Training Loops.}
\label{tab:hyperparameters}
\end{table}

For EBGAN-MDN, the \(\alpha\) parameter was set to 1, and dynamic scaling was disabled (\textit{Dynamic Scaling = False}).
\begin{table}[h!]
\centering
\begin{tabular}{|l|l|c|c|}
\hline
\textbf{Model Component} & \textbf{Layer Type} & \textbf{Output Size} & \textbf{Activation} \\
\hline
\multicolumn{4}{|c|}{\textbf{EBGAN-MDN (Ours)}} \\
\hline
Energy Model & Linear & 64 & ReLU \\
Energy Model & Linear & 64 & ReLU \\
Energy Model & Linear & 1 & - \\
\hline
MDN Generator & Linear & 64 & ReLU \\
MDN Generator & Linear & 64 & ReLU \\
MDN Generator & Mixture Output & $\pi, \mu, \sigma$ (Gaussian Params) & - \\
\hline
\multicolumn{4}{|c|}{\textbf{Explicit BC}} \\
\hline
Explicit BC & Linear & 64 & ReLU \\
Explicit BC & Linear & 64 & ReLU \\
Explicit BC & Linear & 64 & ReLU \\
Explicit BC & Linear & 64 & ReLU \\
Explicit BC & Linear & 1 & - \\
\hline
\multicolumn{4}{|c|}{\textbf{MDN}} \\
\hline
MDN & Linear & 64 & ReLU \\
MDN & Linear & 64 & ReLU \\
MDN & Mixture Output & $\pi, \mu, \sigma$ (Gaussian Params) & - \\
\hline
\multicolumn{4}{|c|}{\textbf{cGAN}} \\
\hline
Generator & Linear & 64 & ReLU \\
Generator & Linear & 64 & ReLU \\
Generator & Linear & 1 & - \\
\hline
Discriminator & Linear & 64 & ReLU \\
Discriminator & Linear & 64 & ReLU \\
Discriminator & Linear & 1 & - \\
\hline
\multicolumn{4}{|c|}{\textbf{EBGAN}} \\
\hline
Energy Model & Linear & 64 & ReLU \\
Energy Model & Linear & 64 & ReLU \\
Energy Model & Linear & 1 & - \\
\hline
Generator & Linear & 64 & ReLU \\
Generator & Linear & 64 & ReLU \\
Generator & Linear & 1 & - \\
\hline
\multicolumn{4}{|c|}{\textbf{IBC}} \\
\hline
IBC & Linear & 64 & ReLU \\
IBC & Linear & 64 & ReLU \\
IBC & Linear & 64 & ReLU \\
IBC & Linear & 1 & - \\
\hline
\end{tabular}
\caption{Architectural details for each model for benchmark 1: hyperbola. Each block corresponds to a specific component, and layers within each component are listed sequentially.}
\label{tab:architectures}
\end{table}

\textbf{Ours (EBGAN-MDN):} The proposed model combines an energy-based network with a Mixture Density Network (MDN). The \textit{Energy Model} is implemented using a MLP to compute energy values for given inputs and actions. It consists of an input layer that concatenates the input condition and action, followed by three fully connected layers with ReLU activations. The output layer produces a single energy value. The \textit{MDN Generator} is responsible for generating samples. It utilizes a sequential structure comprising two hidden layers with ReLU activations, followed by separate output layers for mixture weights, means, and variances. The generator is parameterized to handle isotropic noise and uses temperature scaling to control the sharpness of output distributions.

\textbf{Explicit Behavioral Cloning (BC):} The MLP architecture used in this approach comprises five hidden layers, each with 64 neurons and ReLU activations. The input layer takes the condition, and the output layer predicts the target actions directly using a MSE loss function. This straightforward architecture focuses on directly learning the mapping from inputs to outputs without any latent representation or probabilistic modeling.

\textbf{MDN:} The Mixture Density Network follows a similar architecture to the MDN Generator in the EBGAN-MDN model. It uses two hidden layers with ReLU activations. The network outputs mixture weights, means, and variances for the predefined number of Gaussians. The noise type used is isotropic.

\textbf{Conditional Generative Adversarial Networks (cGAN):} The \textit{Generator} network in cGAN comprises three layers: an input layer that combines the latent input with the condition, followed by two hidden layers with ReLU activations, and an output layer that predicts the target actions. The \textit{Discriminator} network, modified to output raw logits instead of probabilities, follows a similar structure. The input combines the condition with the generated actions, which are passed through two hidden layers with ReLU activations and an output layer producing a single scalar value.

\textbf{Energy-Based Generative Adversarial Networks (EBGAN):} The \textit{Energy Model} in EBGAN is implemented as an MLP with three hidden layers and ReLU activations. The generator architecture mirrors that of the cGAN generator, with an input layer that combines the latent space and condition, followed by two hidden layers, and an output layer predicting the action. The energy model evaluates the compatibility of input and output, forming the basis for training the generator.

\textbf{Implicit Behavioral Cloning (IBC):} The energy-based IBC model is an MLP with four hidden layers. The input layer concatenates the condition and target actions, while the output layer provides a single energy score. ReLU activations are used throughout the network. Unlike explicit BC, IBC does not directly predict actions but instead optimizes energy scores to infer the most likely target for a given condition.

Each architecture is carefully designed to suit the modeling approach, balancing complexity with the capability to learn multimodal mappings. The hidden size is consistently set to 64 neurons per layer to ensure comparability across all models.

\subsubsection{Experimental Setup: Hyperbola} \label{hyperbola_setup}
\begin{figure}[h!]
    \centering
    \includegraphics[width=0.4\linewidth]{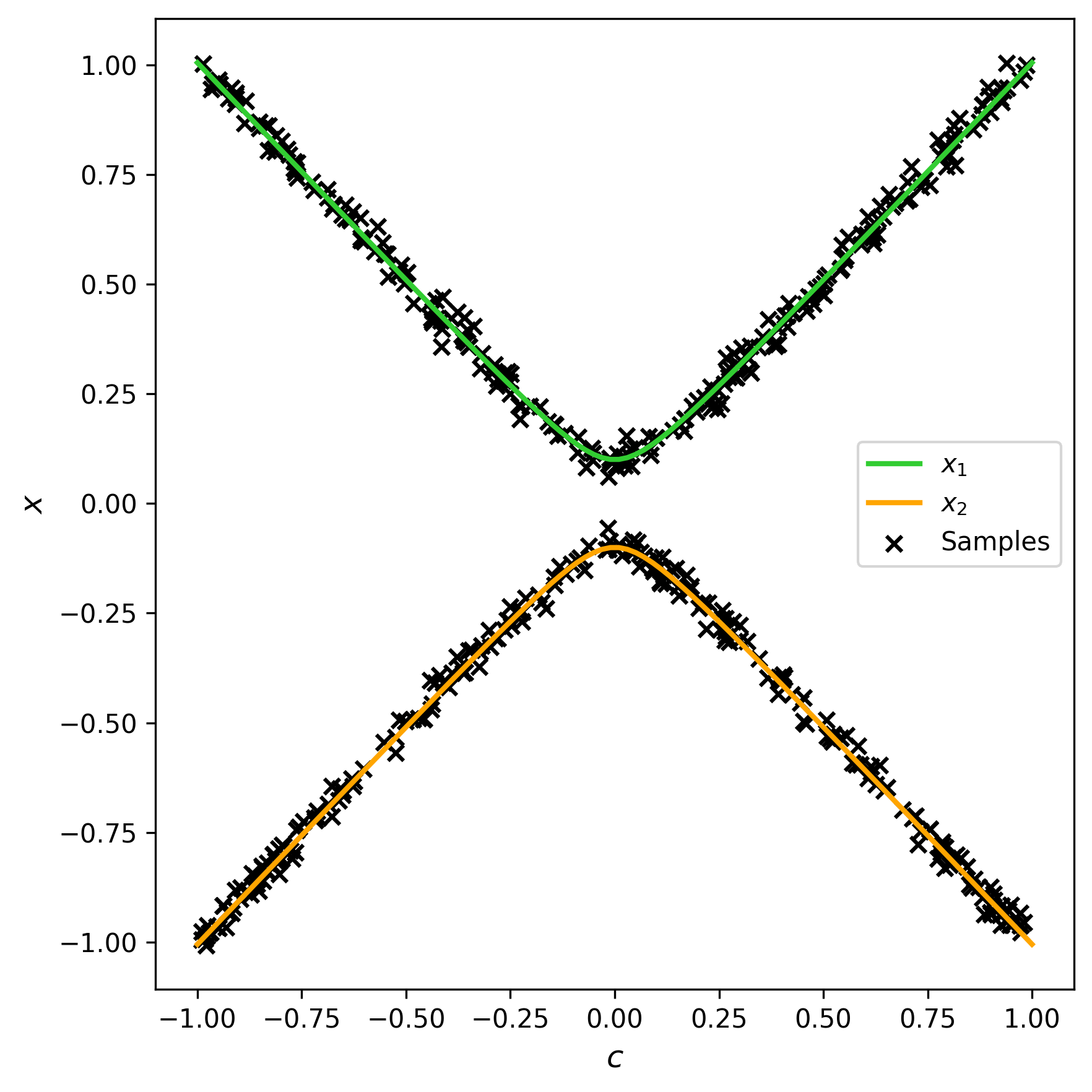}
        \caption{The hyperbola function (test set of our experiment)}
    \label{fig:results_b1_h}
\end{figure}

In this experiment, we evaluate the ability of our approach in comparison to related methods to reconstruct the multi-modal nature of a hyperbolic function under noisy conditions as seen in Figure \ref{fig:results_b1_h}. The hyperbolic function is defined as:
\begin{equation}
    x_1 = \sqrt{a^2 + b^2c^2}, \quad x_2 = -\sqrt{a^2 + b^2c^2},
\end{equation}
where \(c\) serves as the input (in cGAN related frameworks also called condition), and \(x_1, x_2\) are the two possible labels for each \(c\). Therefore this task presents with two conflicting samples -- two modes -- which need to be learned. We will use $a = 0.1$ and $b = 1$ for this experiment.

To simulate realistic conditions, we introduce Gaussian noise into the outputs. For each \(c\), the corresponding \(x_1\) or \(x_2\) is selected randomly, and a small amount of Gaussian noise is added ($\mathcal{N}(0, 0.02)$). This provides a more realistic syntetic dataset.

The dataset is divided into three sets: training, validation, and test. The training set consists of 500 samples with a batch size of 32, while both the validation and global test sets contain 100 samples. The global test set is fixed and consistent across all training runs and models to ensure comparability. The test inputs are uniformly spaced across the range \([-1, 1]\), and their true outputs are computed using the hyperbolic function.

\paragraph{Results}

\begin{table}[h!]
\vskip 0.15in
\begin{center}
\begin{small}
\begin{sc}
\begin{tabular}{lcccc}
\toprule
Model & KL Divergence & Wasserstein Distance & Mode Coverage (\%) & Avg. Modes Captured \\
\midrule
EBGAN-MDN       & $\mathbf{2.11 \pm 0.58}$ & $\mathbf{0.0076 \pm 0.0009}$ & $\mathbf{99.40 \pm 0.80}$ & $\mathbf{1.994 \pm 0.008}$ \\
Explicit BC     & $14.54 \pm 0.30$         & $0.0281 \pm 0.0004$          & $0.00 \pm 0.00$           & $0.032 \pm 0.039$ \\
MDN  & $2.41 \pm 0.63$         & $0.0080 \pm 0.0007$          & $98.80 \pm 1.17$          & $1.988 \pm 0.012$ \\
cGAN             & $4.67 \pm 2.52$         & $0.0119 \pm 0.0042$          & $65.00 \pm 37.64$         & $1.642 \pm 0.390$ \\
EBGAN            & $6.92 \pm 3.40$         & $0.0140 \pm 0.0060$          & $0.00 \pm 0.00$           & $0.990 \pm 0.009$ \\
IBC              & $3.83 \pm 0.34$         & $0.0098 \pm 0.0007$          & $76.80 \pm 3.31$          & $1.768 \pm 0.033$ \\
\bottomrule
\end{tabular}
\end{sc}
\end{small}
\end{center}
\caption{Comparison of Models on Benchmark 1. Metrics include KL Divergence, Wasserstein Distance, Mode Coverage, and Average Modes Captured. Best values are highlighted in bold.}
\label{model-comparison_bench1}
\vskip -0.1in
\end{table}

\begin{figure}[h!]
\begin{center}
\centerline{\includegraphics[width=0.6\linewidth]{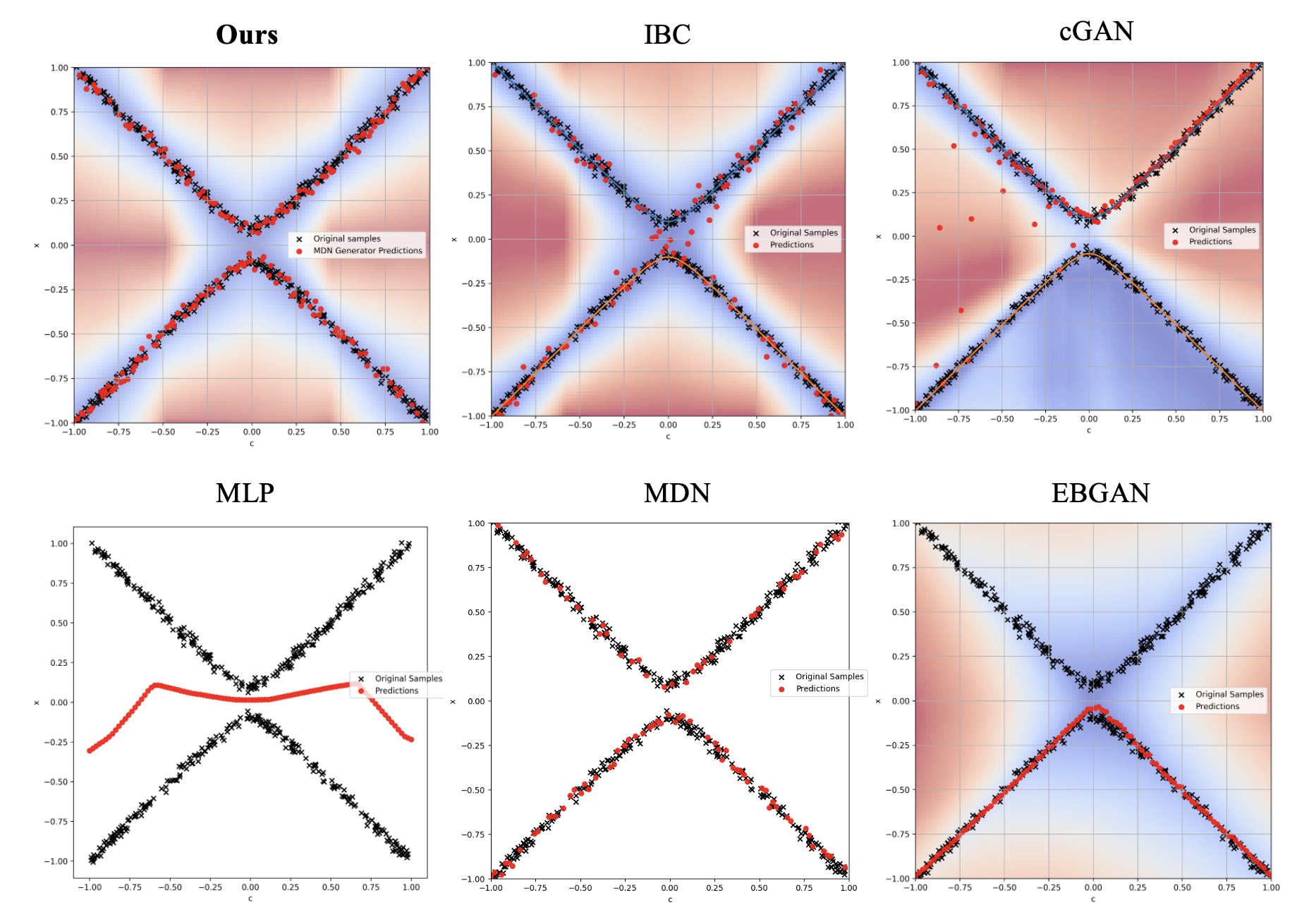}}
\caption{Visualization of model predictions and energy landscapes (or discriminator landscape) on the hyperbola task. Red indicating low plausibility (high energy) and blue indicating high plausibility (low energy).}
\label{ig:App_full_comp_B1}
\end{center}
\vskip -0.3in
\end{figure}

Table \ref{model-comparison_bench1} and Figure \ref{ig:App_full_comp_B1} presents the results from our experiment with the hyperbola. The EBGAN-MDN model achieves the lowest KL divergence (\(2.11 \pm 0.58\)) and Wasserstein distance (\(0.0076 \pm 0.0009\)), along with near-perfect mode coverage (\(99.40\% \pm 0.80\)) and average modes captured (\(1.994 \pm 0.008\)). 

\begin{figure}[h!]
    \centering
    \includegraphics[width=0.8\linewidth]{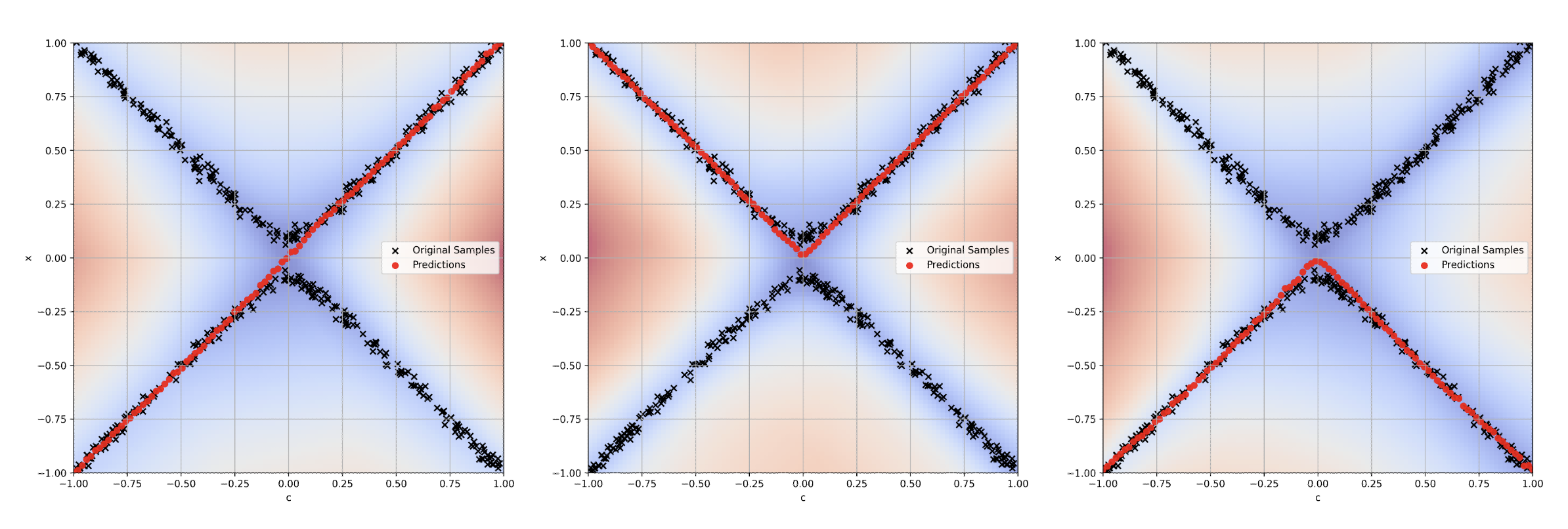}
    \caption{Visualization of different types of mode collapse for EBGAN for Benchmark 1: Hyperbola. Each were observed in the experiment for Benchmark 1: Hyperbola.}
        \label{fig:EBGAN_colage}
\end{figure}

\begin{figure}[h!]
        \centering
        \includegraphics[width=0.8\linewidth]{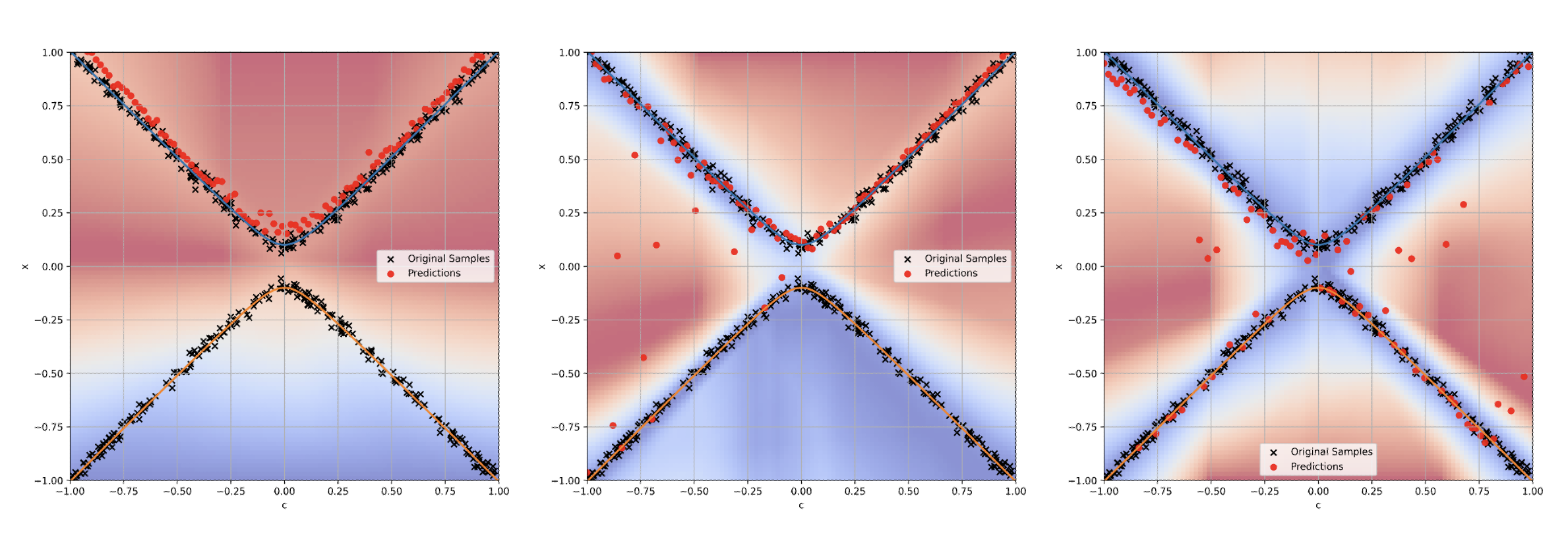}
        \caption{Visualization of different types of partial mode collapse and discriminator landscapes for GAN for Benchmark 1: Hyperbola. The image on the left suffers fully from mode collapse while the other only partially. The images show the variance in GAN training very well: these are three of five results of independent GAN training runs.}
        \label{fig:GAN_colage}
    \end{figure}

In contrast, the explicit behavior cloning (E-BC) model fails entirely, with zero mode coverage and a high KL divergence (\(14.54 \pm 0.30\)), reflecting its inability to handle multimodal tasks with an MSE loss. The cGAN achieves moderate results but suffers from high variance due to mode collapse (\(65.00\% \pm 37.64\) mode coverage), making it inconsistent. This can also be seen in Figure \ref{fig:GAN_colage} in the difference between final results. While some runs produced a two-modal data representation, the model remains unreliable overall. The EBGAN model struggles to represent the distribution, with no runs successfully covering both modes. Its average modes captured (\(0.990 \pm 0.009\)) indicates a tendency to default to capturing only one of the two modes which can be seen in \ref{fig:EBGAN_colage}.

The IBC model performs reasonably well (\(76.80\% \pm 3.31\) mode coverage), but its consistency and accuracy remain below those of EBGAN-MDN. When comparing EBGAN-MDN to MDN with two Gaussians, the former outperforms the latter in all metrics, indicating that the inclusion of an energy model significantly enhances the representation of multimodal distributions.

Overall, EBGAN-MDN emerges as the most effective and reliable approach, consistently capturing the true distribution across all metrics, outperforming other models.

\subsubsection{Experimental Setup: Lines} \label{Line_setup}
In this experiment, we evaluate the ability of our approach in comparison to related methods to reconstruct the three-modal nature of three lines forming a star under noisy conditions. The function is defined as:
\begin{equation}
    x_1 = m \cdot c, \quad x_2 = - m \cdot c, \quad x_3 = 0
\end{equation}
where \(c\) serves as the input, and \(x_1, x_2, x_3\) are the three possible labels for each \(c\). Therefore this task presents with three conflicting samples -- three modes -- which need to be learned. We set $a=1$ for this experiment.

To mimic real-world scenarios, Gaussian noise is introduced into the outputs. For each $c$ one of the three labels are randomly chosen, and a small Gaussian noise term ($\sigma = 0.02$) is added. This creates a more realistic synthetic dataset.

The dataset is split into training, validation, and test sets. The training set contains $500$ samples with a batch size of $32$, while both the validation and test sets consist of $100$ samples. 
To ensure comparability, the global test set is fixed and consistent across all experiments. 
Test inputs are uniformly distributed within the range $[-1,1]$, with their corresponding true 

\begin{figure}[h!]
  \centering
  \includegraphics[width=0.6\textwidth]{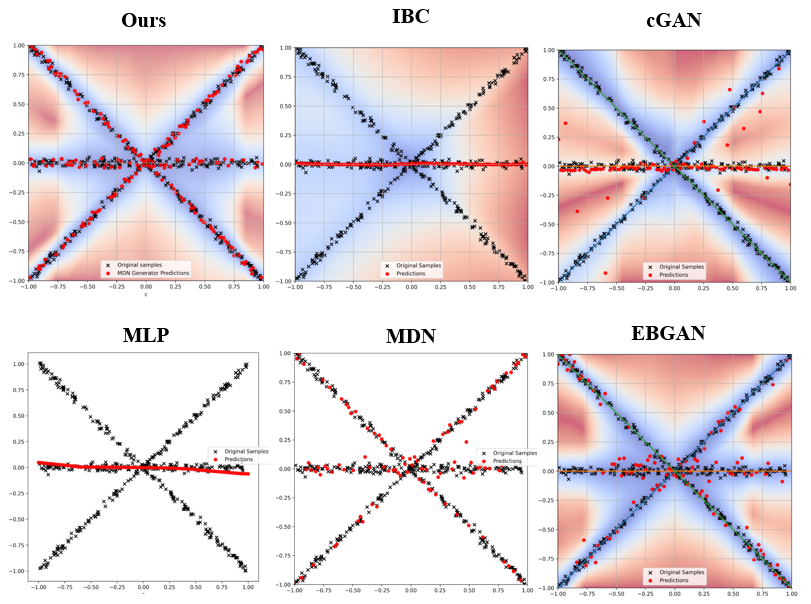}
  \caption{Visualization of model predictions and energy landscapes (or discriminator landscape) on the lines task. Red indicating
low plausibility (high energy) and blue indicating high plausibility
(low energy).}
  \label{fig:lines_bench_all}
\end{figure}

\begin{table}[h!]
\vskip 0.15in
\begin{center}
\begin{small}
\begin{sc}
\begin{tabular}{lcccc}
\toprule
Model        & KL                  & W                  & TMC (\%)                  & AMD                     \\
\midrule
Explicit BC  & $10.4406$          & $0.0242$          & $6.00 \pm 0.00$           & $1.0860$                \\
MDN          & $3.4739$           & $0.0154$          & $\mathbf{95.80 \pm 1.17}$          & $\mathbf{2.9580}$                \\
cGAN         & $8.6680$           & $0.0207$          & $9.20 \pm 2.14$           & $1.4300$                \\
IBC          & $3.7095$           & $0.0182$          & $55.20 \pm 3.87$          & $2.4920$                \\
\midrule
EBGAN        & $10.8500$          & $0.0256$          & $6.40 \pm 0.49$           & $1.1460$                \\
EBGAN-MDN    & $\mathbf{3.0584}$           & $\mathbf{0.0154}$          & $86.40 \pm 16.81$         & $2.8120$                \\
\bottomrule
\end{tabular}
\end{sc}
\end{small}
\end{center}
\caption{Comparison of Models on Benchmark 1: Lines. Metrics include KL Divergence (KL), Wasserstein Distance (W), Total Mode Coverage (\%), and Average Modes Captured (AMD).}
\label{model-comparison-combined}
\vskip -0.1in
\end{table}

\paragraph{Results}

The results in Table~\ref{model-comparison-combined} and Figure~\ref{fig:lines_bench_all} clearly demonstrate the importance of MDNs for handling multi-modal tasks. EBGAN-MDN and MDN outperform all other models in Total Mode Coverage (TMC) and Average Modes Captured (AMC), with MDN achieving 95.80\% TMC and 2.9589 AMC, and EBGAN-MDN achieving 86.40\% TMC and 2.8120 AMC. In contrast, model architectures without MDNs, such as Explicit BC, cGAN, and EBGAN, fail to capture the three-modal structure, as evidenced by their low TMC values ($<10\%$) and poor KL Divergence and Wasserstein Distance metrics. 

The adversarial framework in EBGAN-MDN enhances precision, as reflected in its lowest Wasserstein Distance and KL divergence. However, the MDN baseline achieves higher TMC and AMD, highlighting the inherent strength of explicitly modeling multi-modal distributions. This comparison underscores the need for MDNs in capturing diverse outputs, while adversarial components like EBGAN-MDN improve precision and robustness against noise. Models without MDNs, such as cGAN and IBC, suffer from severe mode collapse and limited coverage, emphasizing the necessity of integrating MDNs for multi-modal learning tasks.

The architecture  information can be taken from the hyperbola experiment and the hyperparameter stay the same except for the number of Gaussian which are increased to 4.

\subsubsection{Additional Benchmarking: Noise Type Experiments}\label{noise}

\paragraph{Experiment Setup}
For these experiments, we used the same EBGAN-MDN parameters and setup as in the hyperbola benchmark described in Section~\ref{hyperbola_setup}. The only changes were the type and level of noise used in the MDN generator. Each configuration was trained and evaluated over five runs, with results reported as mean and standard deviation across these trials. The metrics used for evaluation remain the same: Mode Coverage, Average Modes Captured, KL Divergence, and Wasserstein Distance.


\paragraph{Results}
The results of the noise benchmarking experiments are summarized in Table~\ref{tab:noise_comparison}, which highlights the effects of different noise models on the EBGAN-MDN framework's ability to capture the multi-modal distribution of the hyperbola dataset.

\begin{table}[h!]
\vskip 0.15in
\begin{center}
\begin{small}
\begin{sc}
\begin{tabular}{lcccc}
\toprule
Noise Type & KL Divergence & Wasserstein Distance & Mode Coverage (\%) & Avg. Modes Captured \\
\midrule
Diagonal  & \(1.7141 \pm 0.3794\) & \(0.0073 \pm 0.0008\) & \(98.20 \pm 1.33\) & \(1.982 \pm 0.013\) \\
Isotropic & \(2.3632 \pm 0.8685\) & \(\mathbf{0.0071 \pm 0.0009}\) & \(97.60 \pm 3.32\) & \(1.976 \pm 0.033\) \\
\makecell[l]{Isotropic \\ (Across Clusters)} & \(2.4028 \pm 0.5032\) & \(0.0085 \pm 0.0014\) & \(86.80 \pm 15.98\) & \(1.840 \pm 0.196\) \\
Fixed (\(10^{-3}\)) & \(\mathbf{1.4187 \pm 0.4479}\) & \(0.0073 \pm 0.0004\) & \(99.20 \pm 1.60\) & \(1.992 \pm 0.016\) \\
Fixed (\(10^{-2}\)) & \(1.6183 \pm 0.5117\) & \(0.0074 \pm 0.0008\) & \(\mathbf{99.60 \pm 0.49}\) & \(\mathbf{1.996 \pm 0.005}\) \\
Fixed (\(10^{-1}\)) & \(2.6629 \pm 0.3755\) & \(0.0079 \pm 0.0017\) & \(90.40 \pm 3.44\) & \(1.904 \pm 0.034\) \\
\makecell[l]{Laplace \\ (Diagonal)} & \(1.4938 \pm 0.2573\) & \(\mathbf{0.0071 \pm 0.0008}\) & \(99.40 \pm 0.49\) & \(1.994 \pm 0.005\) \\
\bottomrule
\end{tabular}
\end{sc}
\end{small}
\end{center}
\caption{Performance of EBGAN-MDN under different noise assumptions for the MDN generator on the hyperbola benchmark. KL Divergence and Wasserstein Distance indicate distribution alignment, whereas Mode Coverage and Average Modes Captured evaluate how thoroughly all modes are represented.}
\label{tab:noise_comparison}
\end{table}

\begin{itemize}
    \item \textbf{Fixed Noise:} Both fixed noise levels at \(10^{-2}\) and \(10^{-3}\) yield excellent mode coverage (\(>99\%\)) and tight KL Divergence, with \(10^{-2}\) achieving the best overall results. This highlights the stability of fixed noise in modeling local variations, although it might be less flexible for datasets with unknown noise properties.
    \item \textbf{Diagonal and Isotropic Noise:} These assumptions perform well, with mode coverage above \(97\%\) and good KL Divergence. However, they sometimes underperform fixed noise, possibly due to their inability to precisely model noise distribution around each mode.
    \item \textbf{Laplace Noise:} Laplace noise demonstrates strong performance across all metrics. Its heavier tails aid in capturing outliers near each mode, contributing to consistent high mode coverage and distribution alignment.
    \item \textbf{Isotropic Across Clusters:} This configuration suffers from reduced mode separation, leading to lower mode coverage and higher variability. This indicates the challenges of applying a uniform isotropic noise across multi-modal distributions.
\end{itemize}

These results emphasize the robustness of fixed noise and Laplace noise under our specific data generation process, while also highlighting the potential trade-offs in flexibility and adaptability for other settings.


\subsubsection{Additional Benchmarking: Generator Configuration Experiments}\label{experiment_gen}
\paragraph{Experimental Configurations} To explore the implications of incorporating generator samples into the InfoNCE objective, we evaluated three experimental setups:
\begin{itemize}
    \item \textbf{Removing the Generator Term:} In this configuration, the denominator only includes negative samples drawn from the data distribution $p(x)$. This tests whether the energy model can effectively learn the landscape without any influence from the generator. This recovers the InfoNCE Loss from \cite{florence2022implicit}:
    \[
    \mathcal{L}_{\text{InfoNCE}}(\theta, \phi) = -  \sum_{i=1}^D
    \log \frac{\exp(-E_\theta(c_i, x_i))}{\exp(-E_\theta(c_i, x_i)) + \sum_{j=1}^{M} \exp(-E_\theta(c_i, x_i^j))}
    \]
    \item \textbf{Standard Generator Inclusion:} One generator sample was included in the denominator as described in the extended InfoNCE objective. This approach ensures the energy model learns to recognize valid pairs while being exposed to the generator's evolving outputs. 
    \[
    \mathcal{L}_{\text{EBM}}(\theta, \phi) = -  \sum_{i=1}^D
    \log \frac{\exp(-E_\theta(c_i, x_i))}{\exp(-E_\theta(c_i, x_i)) + \sum_{j=1}^{M} \exp(-E_\theta(c_i, x_i^j)) + \exp(-E_\theta(c_i, G_\phi(z_i \mid c_i)))}
    \]
    \item \textbf{Equal Ratio Inclusion:} N-1 generator samples were included in the denominator as described in the extended InfoNCE objective.
    \[
    \mathcal{L}_{\text{EBM}}(\theta, \phi) = -  \sum_{i=1}^D
    \log \frac{\exp(-E_\theta(c_i, x_i))}{\exp(-E_\theta(c_i, x_i)) + \sum_{j=1}^{M} \exp(-E_\theta(c_i, x_i^j)) + \sum_{k=1}^{M}\exp(-E_\theta(c_i, G_\phi(z_k \mid c_i)))}
    \]
    \item \textbf{Dynamic Scaling:} To address the tension introduced by including valid generator samples in the denominator, we experimented with dynamic scaling. In this configuration, the contribution of the generator term to the denominator was decreased over the training period. We additonally set $N=1$ as in the standard gernator inclusion and only added one genertor sample. For example: 
    \begin{align}
        \alpha_t = \max\{1 - \frac{t}{T}, a_{\min}\}
    \end{align}
    where $\alpha_t$ is the factor that is multiplied with the generator term in the denominator, $t$ is the epoch, $T$ is the total number of epochs, and $a_{\min}$ is a minimum for ensuring that the generator sample's weight is never 0 (We used $0.1$ in our experiments). This ensured that the generator's influence diminished as its outputs became more valid, preventing the unintended penalization of plausible samples.
    \[
    \mathcal{L}_{\text{EBM}}(\theta, \phi) = -  \sum_{i=1}^D
    \log \frac{\exp(-E_\theta(c_i, x_i))}{\exp(-E_\theta(c_i, x_i)) + \sum_{j=1}^{M} \exp(-E_\theta(c_i, x_i^j)) + \alpha_t\exp(-E_\theta(c_i, G_\phi(z_i \mid c_i)))}
    \]
\end{itemize}

We use Benchmark 1 and the hyperbola experiment to investigate the configurations. Accordingly the implementation details were the same as in \ref{Implementation_app}. We use the same metrics where a lower KL Divergence and Wasserstein Distance indicate better distribution alignment, while higher Total Mode Coverage and Average Modes Covered reflect better mode diversity.

\begin{figure}[h!]
    \centering
    \includegraphics[width=1\linewidth]{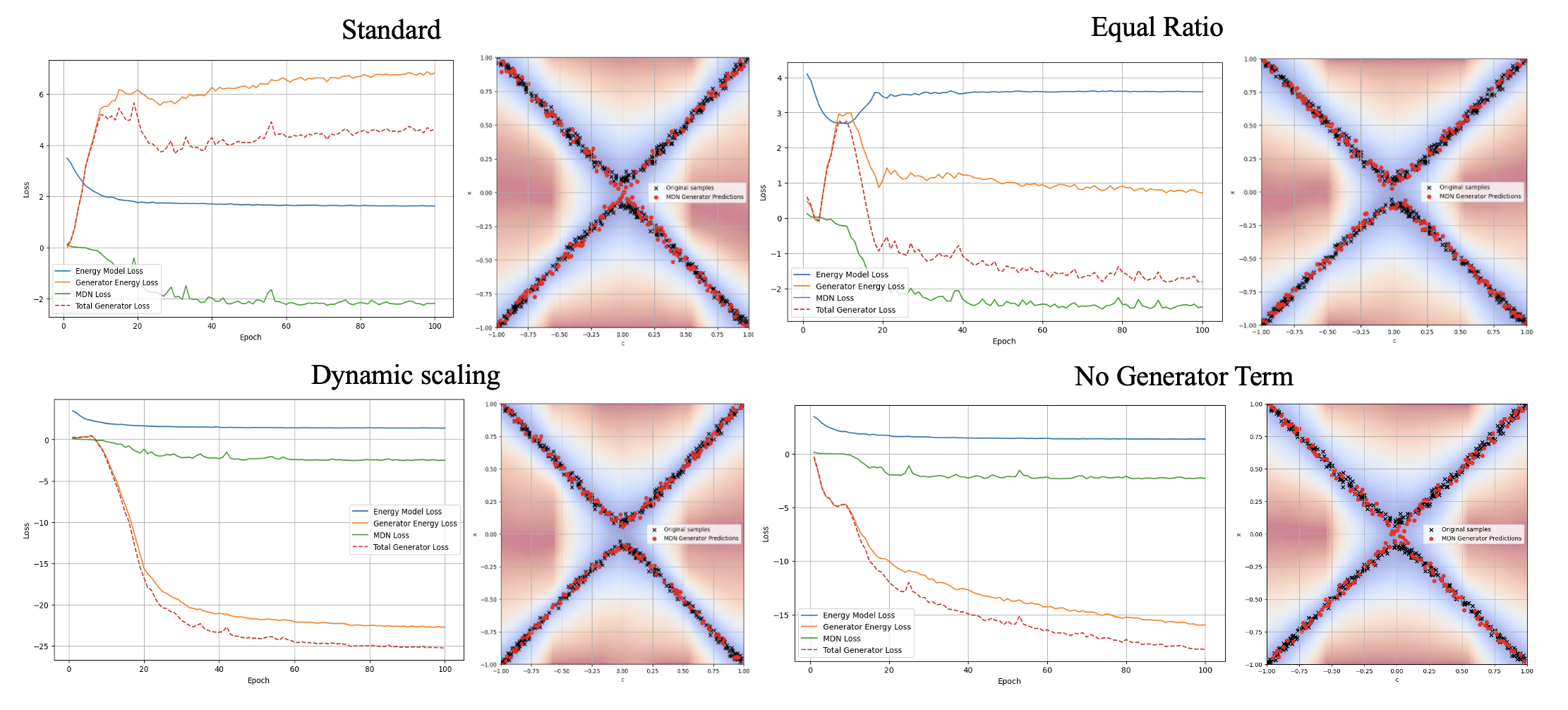}
    \caption{Comparison of loss functions and predictions across different configurations of generator inclusion. (Top Left) "Standard Generator Inclusion" shows a more extreme example recorded during the experiment with strong oscillations in the generator energy loss. (Top Right) "Equal Ratio Inclusion" includes multiple generator samples, resulting in more diverse predictions but introducing instability. (Bottom Left) "Dynamic Scaling" smooths the generator energy loss and stabilizes training by reducing the generator term’s influence over time. (Bottom Right) "No Generator Term" eliminates the generator contribution and is therefore not affected by the conflicting objectives. The accompanying hyperbola predictions demonstrate the effect of each configuration on distribution alignment and mode diversity.}
    \label{fig:loss_curves_no_scaling}
\end{figure}

\paragraph{Results} \label{Results_gen_config}The results of the generator configurations are summarized in Table~\ref{tab:results_EP_}, and the training dynamics are illustrated in Figure~\ref{fig:loss_curves_no_scaling}. 
\begin{table}[h!]
\vskip 0.15in
\begin{center}
\begin{small}
\begin{sc}
\begin{tabular}{lcccc}
\toprule
Configuration & KL & WD & TMC (\%) & AMC \\
\midrule
No Generator & $2.2387 \pm 0.7505$ & $0.0076 \pm 0.0010$ & $92.20 \pm 15.60$ & $1.9220 \pm 0.1560$ \\
Standard Generator Inclusion & $1.8267 \pm 0.5521$ & $0.0070 \pm 0.0016$ & $\mathbf{98.00 \pm 0.90}$ & $\mathbf{1.9800 \pm 0.0089}$ \\
Equal Ratio Inclusion & $1.8928 \pm 0.8036$ & $0.0072 \pm 0.0011$ & $97.80 \pm  2.86$ & $1.9780 \pm 0.0286$ \\
Dynamic Scaling & $\mathbf{1.6934 \pm 0.4217}$ & $\mathbf{0.0066 \pm 0.0006}$ & $96.20 \pm 5.64$ & $1.9600 \pm 0.0559$ \\
\bottomrule
\end{tabular}
\end{sc}
\end{small}
\end{center}
\caption{Comparison of generator configurations. Metrics include KL Divergence (KL), Wasserstein Distance (W), Total Mode Coverage (\%), and Average Modes Captured (AMD). Lower KL Divergence and Wasserstein Distance indicate better distribution alignment, while higher Mode Coverage and Average Modes Covered reflect better mode diversity.}
\label{tab:results_EP_}
\vskip -0.1in
\end{table}

\textbf{Impact of Generator Inclusion:}
The inclusion of generator samples ("Standard Generator Inclusion" and "Equal Ratio Inclusion") improved the energy model’s ability to adapt to the evolving generator distribution compared to the "No Generator" configuration. While "No Generator" achieved reasonable TMC (92.20\%) and AMC (1.92), it resulted in significantly higher KL Divergence (2.24), indicating poorer generalization and weaker distribution alignment.

"Equal Ratio Inclusion" further improved Mode Coverage to 97.80\% by including multiple generator samples, but at the cost of introducing training instability, as reflected in higher KL variance and oscillatory loss curves in Figure~\ref{fig:loss_curves_no_scaling}. On the other hand, "Standard Generator Inclusion" provided the most consistent results across metrics, achieving the highest TMC (98.00\%) and AMC (1.98), while maintaining competitive KL Divergence (1.83) and WD (0.0070).

\textbf{Impact of Dynamic Scaling on Stability:}
The inclusion of dynamic scaling improved training stability significantly. As shown in Figure~\ref{fig:loss_curves_no_scaling}, the generator energy loss stabilized with dynamic scaling, showing a smooth and steady decrease compared to the oscillatory behavior in "Standard Generator Inclusion" and "Equal Ratio Inclusion." This stability stems from gradually reducing the influence of generator samples as their outputs became more plausible, mitigating the tension between the energy-based model and the generator.

Dynamic scaling also achieved the lowest KL Divergence (1.69) and WD (0.0066), indicating superior distribution alignment. However, it slightly reduced Mode Coverage (96.20\%) and AMC (1.96) compared to "Standard Generator Inclusion." These results suggest that dynamic scaling is particularly beneficial in scenarios where training stability is critical, even if it introduces a minor trade-off in capturing mode diversity. For tasks requiring high stability or consistency, dynamic scaling provides a robust solution.

\subsection{Benchmark Robot tasks}

\subsubsection{Robot Tasks}
\textbf{IK 2 link Robot:} consists of a 2-link robotic arm (both length 3m) with two possible configurations to reach a target position (tolerance: 0.2 m). The task is considered successful if the robot reaches the target position with either of the two valid configurations. The data consists of 512 
data points generated by solving inverse kinematics using optimization. The action is the angle of the two links (2D), and the observation is the position of the end effector (2D).

\textbf{IK UR5 Robot:} consists of a 6-DOF UR5 robotic arm with multiple valid joint configurations to reach a target position (tolerance: 10 cm). The task is considered successful if the robot reaches the target position with any valid joint configuration. The data consists of 3,000 data points generated by solving inverse kinematics using PyBullet inverse kinematics solver. 
The action is the joint angles of the UR5 robot (6D), and the observation is the position of the end effector (2D).

\textbf{Push-T:} consists of a circle pusher and a T-shaped block simulated in PyBullet. The goal is to push the T-shaped block to a target position. The task is considered successful if the block reaches the target position within a certain threshold. 
The data is generated using oracl method in simulartion simialr to \cite{florence2022implicit}. The action is the 
change of position of the circle pusher (2D) and the observation is the keypoints of the T-shaped block (20D).

\textbf{Block Push:} consists of a simulated 6DoF robot in PyBullet equipped with a small cylindrical end effector to push the block to two zones(multistage). The success is defined as the block being within a small threshold of the target zone. The task is considered successful if the block reaches the target zone within a certain threshold. 
The data consists of 2,000 demonstrations from \cite{chi2023diffusion}, with actions representing changes in end effector position and orientation (2D), and observations including block translation (2D), block orientation (1D), effector translation (2D), effector target translation (2D), target translation (2D), and target orientation (1D).

\subsubsection{Implementation Details}

The specific training hyperparameters for each model are summarized in Table~\ref{tab:hyperparameters}, while Table~\ref{tab:architectures} describes the architectural details. 

Note that for 2 link inverse kinematics, the hidden layer is of size 64 and of 2 layers. The discriminator and energy model are improved by 5 iterations per 1 generator iteration.

\begin{table}[h!]
\centering
\begin{tabular}{|l|c|c|c|c|c|c|c|}
\hline
\textbf{Model} & \textbf{Gaussians} &  \textbf{Hidden Size} & \textbf{Epochs} & \textbf{Batch Size} & \textbf{LR} & \textbf{LR-E/D} & \textbf{Negative Samples} \\
\hline
EBGAN-MDN & 10 & 128 & 100 & 32 & 0.0005 & 0.001 & 256 \\
\hline
Explicit BC &  - & 128 & 100 & 32 & 0.001 & - & - \\
\hline
MDN & 10 &  128 & 100 & 32 &  0.001 & - & - \\
\hline
cGAN & - &  128 & 100 & 32 &  0.0005 & 0.001 & - \\
\hline
EBGAN & - &  128 & 100 & 32 & 0.0005 & 0.001 & 256 \\
\hline
IBC & - &  128 & 100 & 32 & - & 0.001 & 256 \\
\hline
\end{tabular}
\caption{Hyperparameters for each model. LR and LR-E/D are the learning rates where LR-E/D captures learning rates of discriminators and Energy models and LR typically for any other used model.}
\label{tab:hyperparameters}
\end{table}

\begin{table}[h!]
\centering
\begin{tabular}{|l|l|c|c|}
\hline
\textbf{Model Component} & \textbf{Layer Type (Repeated)} & \textbf{Output Size} & \textbf{Activation} \\
\hline
\multicolumn{4}{|c|}{\textbf{EBGAN-MDN (Ours)}} \\
\hline
Energy Model & Linear (4 layers) & 128 & ReLU \\
Energy Model & Linear (1 layer) & 1 & - \\
MDN Generator & Linear (4 layers) & 128 & ReLU \\
MDN Generator & Mixture Output & $\pi, \mu, \sigma$ (Gaussian Params) & - \\
\hline
\multicolumn{4}{|c|}{\textbf{Explicit BC}} \\
\hline
Explicit BC & Linear (4 layers) & 128 & ReLU \\
Explicit BC & Linear (1 layer) & act\_dim & - \\
\hline
\multicolumn{4}{|c|}{\textbf{MDN}} \\
\hline
MDN & Linear (4 layers) & 128 & ReLU \\
MDN & Mixture Output & $\pi, \mu, \sigma$ (Gaussian Params) & - \\
\hline
\multicolumn{4}{|c|}{\textbf{cGAN}} \\
\hline
Generator & Linear (4 layers) & 128 & LeakyReLU \\
Generator & Linear (1 layer) & act\_dim & - \\
Discriminator & Linear (4 layers) & 128 & LeakyReLU \\
Discriminator & Linear (1 layer) & 1 & - \\
\hline
\multicolumn{4}{|c|}{\textbf{EBGAN}} \\
\hline
Energy Model & Linear (4 layers) & 128 & ReLU \\
Energy Model & Linear (1 layer) & 1 & - \\
Generator & Linear (4 layers) & 128 & ReLU \\
Generator & Linear (1 layer) & act\_dim & - \\
\hline
\multicolumn{4}{|c|}{\textbf{IBC}} \\
\hline
IBC & Linear (4 layers) & 128 & ReLU \\
IBC & Linear (1 layer) & 1 & - \\
\hline
\end{tabular}
\caption{Architectural details for each model for benchmark 2. Each block summarizes repeated layers and their activation functions.}
\label{tab:architectures}
\end{table}

\clearpage

\end{document}